\documentclass[manuscript, screen]{jair}

\setcopyright{cc}
\acmDOI{}

\JAIRAE{}
\JAIRTrack{} 
\acmYear{2026}

\RequirePackage[
  datamodel=acmdatamodel,
  style=acmauthoryear,
  backend=biber,
  giveninits=true,
  uniquename=init
  ]{biblatex}
  
\usepackage[ruled,linesnumbered]{algorithm2e}

\begin{document}

\title{Towards Provably Fair Machine Learning: Bayesian Approaches For
Consistent and Transparent Predictions}

\author{Owen O'Neill}
\authornote{Corresponding Author.}
\orcid{0009-0009-1406-9993}
\email{owen.o-neill.1@ucdconnect.ie}
\affiliation{%
  \institution{University College Dublin}
  \city{Dublin}
  \state{Co. Dublin}
  \country{Ireland}
}

\author{Fintan Costello}
\orcid{0000-0002-3953-7863}
\email{fintan.costello@ucd.ie}
\affiliation{%
  \institution{University College Dublin}
  \city{Dublin}
  \state{Co. Dublin}
  \country{Ireland}
}

\renewcommand{\shortauthors}{O'Neill \& Costello}


\begin{abstract}
{\bf Background:}
Machine learning classifiers deployed in high-stakes domains
produce predictions whose quality varies systematically across
subgroups. For granular subgroups defined by intersections of
multiple features, predictions are often inconsistent
with the observed data: the model's outputs contradict the evidence
available for that subgroup. This problem is exacerbated by
regularisation, which improves aggregate performance by collapsing
small subgroups into larger groups, disproportionately affecting
demographic minorities.

{\bf Objectives:}
We aim to formalise what it means for a prediction to be
statistically consistent with the observed data, develop a
classifier that enforces this consistency exhaustively across
every possible subgroup in a categorical dataset, and identify
cases where no consistent prediction is possible.

{\bf Methods:}
We define two requirements for consistent prediction: determinism
(identical individuals receive identical predictions) and
statistical consistency (we cannot reject, at significance level 
$\alpha$, the hypothesis that the predictions for a subgroup were 
drawn from the Bayesian optimal target distribution inferred for 
that subgroup). From these requirements we derive the Fair Bayesian 
classifier, which enforces both across every group and subgroup 
in the data simultaneously and abstains whenever no consistent 
deterministic prediction is possible.

{\bf Results:}
On three benchmark datasets (Adult, COMPAS, and Bank Marketing),
standard classifiers including a decision tree, a neural network, and
a proportional multicalibration post-processor produce
statistically inconsistent predictions for a substantial proportion
of subgroups. The Fair Bayesian classifier achieves zero consistency
error by construction while exceeding baseline accuracy on every
dataset tested, on the subgroups where it predicts. Competitive performance on
the multicalibration metric, against models optimised explicitly
for it, also emerges as a by-product of consistency.

{\bf Conclusions:}
Statistical consistency provides a principled foundation for prediction 
quality with direct implications for algorithmic fairness. Minority 
demographics are disproportionately concentrated in small subgroups, 
and small subgroups are precisely where frequentist inference is least 
reliable; addressing this inference problem is therefore a necessary 
step toward fair ML. By enforcing Bayesian consistency at the finest 
resolution the data supports, the Fair Bayesian classifier demonstrates 
that exhaustive subgroup fairness with principled abstention is achievable 
in practice.
\end{abstract}




\maketitle

\section{Introduction}

Machine learning (ML) models make predictions by generalising from 
training data to new instances. The quality of these predictions 
depends on the evidence behind them: the subset of training data 
that the model treats as most relevant to the input. When this 
subset is large, the observed evidence supports reliable inference 
about the group's relationship with the target variable at the 
population level; when it is small, it does not. In high-stakes 
domains such as finance, criminal justice, and healthcare, where 
algorithmic decisions can significantly impact people's lives, the 
reliability of this inference process is paramount.

Standard frequentist approaches treat observed sample proportions as
reliable estimates of the true underlying probability, regardless 
of sample size. This assumption weakens for small samples, where
there may be insufficient evidence to draw reliable conclusions about the
population. Cunningham et al.\ \cite{cunningham2020} identify a
related pattern: common ML models systematically
underestimate the target rate for all groups in the data, with the
effect being greater for infrequent minority groups. The bias does
not depend on the minority group being a protected demographic; many
relatively rare groups in the dataset are affected. 
This problem is compounded by regularisation, the standard ML 
response to small-sample overfitting, which improves generalisation 
on average by merging small subgroups into larger aggregates. In 
doing so it discards information about the very subgroups whose 
prediction quality is most at risk.

These inference issues have consequences for fair decision making. ML models
deployed in areas such as criminal justice \cite{compas2016},
recruiting \cite{amazon}, and healthcare \cite{obermeyer2019dissecting}
have been shown to produce outcomes that disproportionately
disadvantage minority groups. A significant body of literature has
emerged in response, proposing techniques that modify existing
algorithms (before training, during optimisation, or after
prediction) to satisfy a chosen fairness metric
\cite{alves2023survey}. However, these approaches face persistent
limitations. The fairness metrics most frequently invoked (group
fairness, individual fairness, causal fairness) are often mutually
incompatible: a classifier certified as fair under one metric can
violate another in a provable sense \cite{mehrabibias}. Global
performance metrics, whether fairness metrics defined on broad
demographic groups or aggregate accuracy, obscure harms at the
subgroup level. A model may satisfy a fairness criterion for the
group `women' yet fail for the subgroup `low-income, rural white
women'. Multicalibration and multiaccuracy
\cite{hebert2018} aim to enforce fairness across more subgroups, but
the search space grows exponentially and implementations seldom
consider more than two or three protected variables. A separate 
strand of work applies Bayesian ideas to fairness 
\cite{dimitrakakis2019bayesian, foulds2020bayesian}, typically by 
placing priors over model parameters or subgroup-level rates 
rather than by enforcing consistency between predictions and 
observed evidence.

A further gap concerns abstaining from prediction. No classifier 
can be reliable on every possible input, so the ability to 
decline to predict is a basic safeguard: a model that flags 
the cases it cannot answer is preferable to one forced to 
predict on every input. Abstention is justified on 
two grounds. The first, familiar from the selective prediction 
literature \cite{chow1970optimum, herbei2006classification, 
elyaniv2010selective, wiener2015agnostic}, is insufficient evidence: the data 
available for an individual or subgroup is too sparse to 
support a confident prediction. The second has received far 
less attention: the evidence is strong enough to rule out 
every possible prediction. Consider a group of 100 individuals 
with identical values across every recorded variable, exactly 
50 of whom exhibit the target attribute. The group is large 
enough to be confident that the true target rate is close to 
50\%; that very confidence is what makes neither 
deterministic prediction statistically defensible. 
Predicting the entire group positive would knowingly 
misclassify half; predicting the entire group negative would 
do the same. Current ML methods, including those 
explicitly designed for fairness, provide no mechanism for 
identifying these cases and instead issue a prediction 
regardless.

Both problems (unreliable inference for small subgroups and the 
absence of principled abstention) can be seen as arising from a 
shared root cause: 
standard ML models optimise for aggregate accuracy without 
enforcing statistical consistency between their predictions and 
the evidence for each subgroup. In this paper, we address that 
root cause directly. Rather than training a model to learn 
associations
between features and outcomes and then retrofitting fairness
constraints, we treat classification as a question of statistical
justification. We propose two requirements for any prediction to be
considered consistent: first, identical individuals must receive
identical predictions (determinism); second, the prediction assigned
to every group and subgroup in the data must be statistically
consistent with the target distribution inferred from the observed
sample, assessed via hypothesis test at a chosen significance level
$\alpha$. From these requirements, we derive and implement the Fair
Bayesian (FB) classifier. Unlike conventional ML models, which 
optimise for accuracy while neglecting the reliability of inference 
for small subgroups, the FB classifier is built from the ground up 
around the notion of statistical consistency. It does not modify 
an existing model's outputs, but constructs predictions directly 
from the statistical evidence available for each subgroup. Our 
contributions are as follows:

\begin{enumerate}
    \item A framework for prediction consistency based on Bayesian
    inference. For every subgroup in a categorical dataset, we derive
    the posterior distribution over the unknown target probability and
    test whether the model's predictions are consistent with this
    distribution. This is enforced exhaustively across all subgroups
    defined by any combination of feature values, not just a
    pre-specified set of protected groups.

    \item A classifier that guarantees consistency with principled
    abstention. The FB classifier assigns a deterministic prediction
    to each subgroup where the data supports one, and abstains where
    no deterministic prediction is consistent with the evidence. This
    abstention is not based on model confidence but follows directly
    from the consistency requirement: when the hypothesis test rules
    out both possible predictions, the model must decline to predict.

    \item Empirical demonstration that standard ML models produce
    statistically inconsistent predictions. On three benchmark
    datasets (Adult, COMPAS, and Bank Marketing), we show that a
    decision tree, a neural network, and PMCBoost 
    \cite{lacava2023pmc} all make predictions that contradict the
    statistical evidence for a substantial proportion of subgroups.
    The FB classifier achieves zero consistency error by construction,
    while exceeding baseline accuracy on every dataset tested, on the
    subgroups where it makes predictions. Multicalibration competitive
    with models optimised explicitly for it emerges as a by-product of
    consistency.
\end{enumerate}

The remainder of the paper is organised as follows.
Section~\ref{sec:background} reviews the small-sample inference
problem, the subgroup structure of ML datasets, and the most closely
related prior work. Section~\ref{sec:preliminaries} presents the
Bayesian tools our framework builds on. Section~\ref{sec:framework}
formalises our consistency requirements and derives the inference
framework, including the abstention mechanism.
Section~\ref{sec:algorithm} describes the implementation of the
framework as a practical algorithm.
Section~\ref{sec:experiments} presents our empirical evaluation, and
Section~\ref{sec:discussion} discusses the implications and
limitations of our approach.

\section{Background and Related Work}
\label{sec:background}
\subsection{Small-Sample Inference in Machine Learning}

The introduction highlighted that frequentist inference can be 
unreliable for small samples. Consider the following extreme example. 
A sample of zero positive cases among three individuals yields the 
same observed rate (0\%) as zero positives among one hundred, yet 
the inference we should draw from these samples differs 
substantially. We can infer with reasonable confidence that the 
target is rare in the larger sample; we have too little evidence 
to state the same for the smaller sample. This is not reflected in 
frequentist inference, where the observed rate is assumed to 
directly reflect the true probability regardless of sample size.

Bayesian inference addresses this problem by accounting for the 
increased uncertainty associated with small samples. Given a sample 
of $N$ items containing $K$ instances of $T$ and a uniform prior
$\text{Beta}(1,1)$ over the unknown population probability, the
posterior mean is $(K+1)/(N+2)$: the Rule of Succession
\cite{definetti1937, zabell1989}. This estimate is regressive toward
0.5, with the degree of regression increasing as $N$ falls. Two
groups that share the same observed target rate but differ in size
will therefore receive different population probability estimates,
reflecting the greater uncertainty associated with smaller samples.
This difference is not a bias introduced by the method; it is the
rational consequence of having less information about the smaller
group.

\begin{figure}
    \centering
    \includegraphics[width=0.8\textwidth]{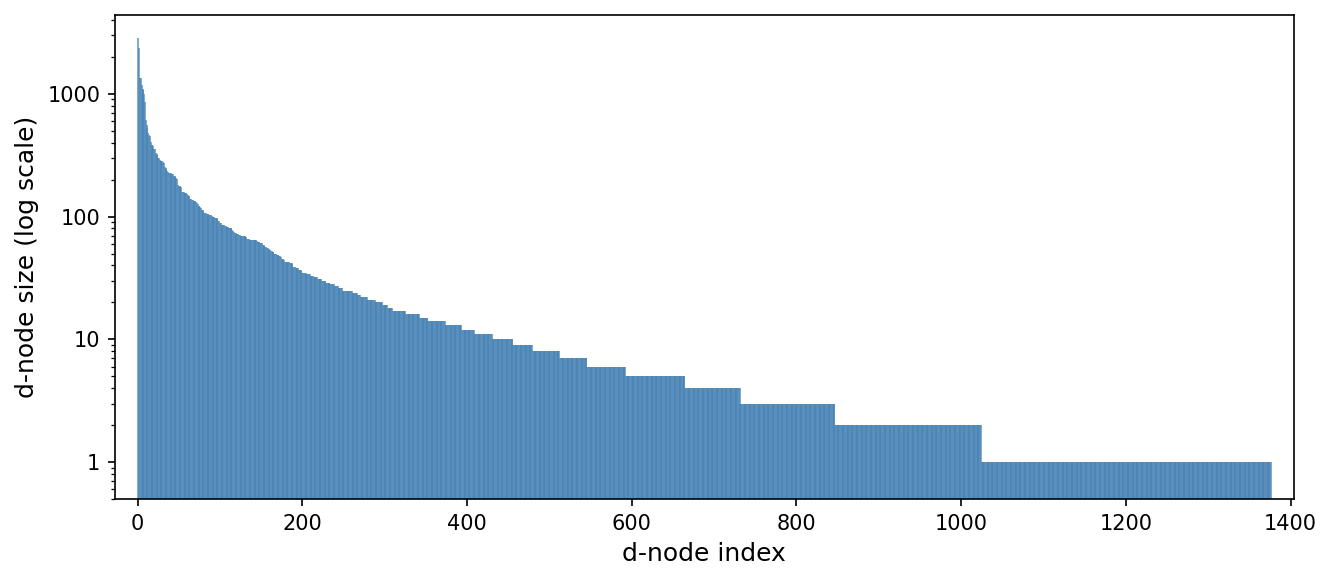}
    \caption{$d$~node Size Distributions of Adult (Log Scale y-axis)}
    \Description{Bar chart showing the distribution of d node sizes for the Adult dataset, displayed with a logarithmic scale on the y-axis.}
    \label{fig:dnodesadult}
\end{figure}

The relevance of this observation to ML-scale inference is not 
immediately obvious. Datasets routinely contain tens of thousands 
of data points, and it might appear that the small-sample scenario 
described above is irrelevant at this scale. Even in these large 
datasets, however, when we view the data at the highest resolution 
available, small subgroups emerge. In categorical data, we can 
consider the level of groups of individuals who share 
identical values across every variable: the most 
informative subgroups 
the data can support. We refer to these groups as $d$~nodes 
(formally defined in Section~\ref{sec:notation}). Across three 
benchmark datasets (Adult, COMPAS, and Bank Marketing), the 
majority of $d$~nodes contain very few individuals. 
Figure~\ref{fig:dnodesadult} shows the $d$~node size distribution 
for the Adult dataset, where most contain fewer than 10 individuals.

While viewing the data from this perspective offers the most 
information, it is typically avoided in ML due to the risks of 
overfitting. Considering the patterns observed in these small 
groups from a frequentist inference perspective produces the 
erroneous predictions as described above. ML therefore applies 
regularisation. By pruning decision tree branches or focusing on 
coarser patterns in the data, the algorithms consolidate these 
small groups, producing more stable predictions. This improves 
generalisation because predictions are averaged across broader 
groups with more data behind them. From the standpoint of 
prediction error, this is effective: variance is reduced at the 
cost of some bias, a trade-off that usually yields better 
out-of-sample performance.

\begin{figure}
    \centering
    \includegraphics[width=0.8\textwidth]{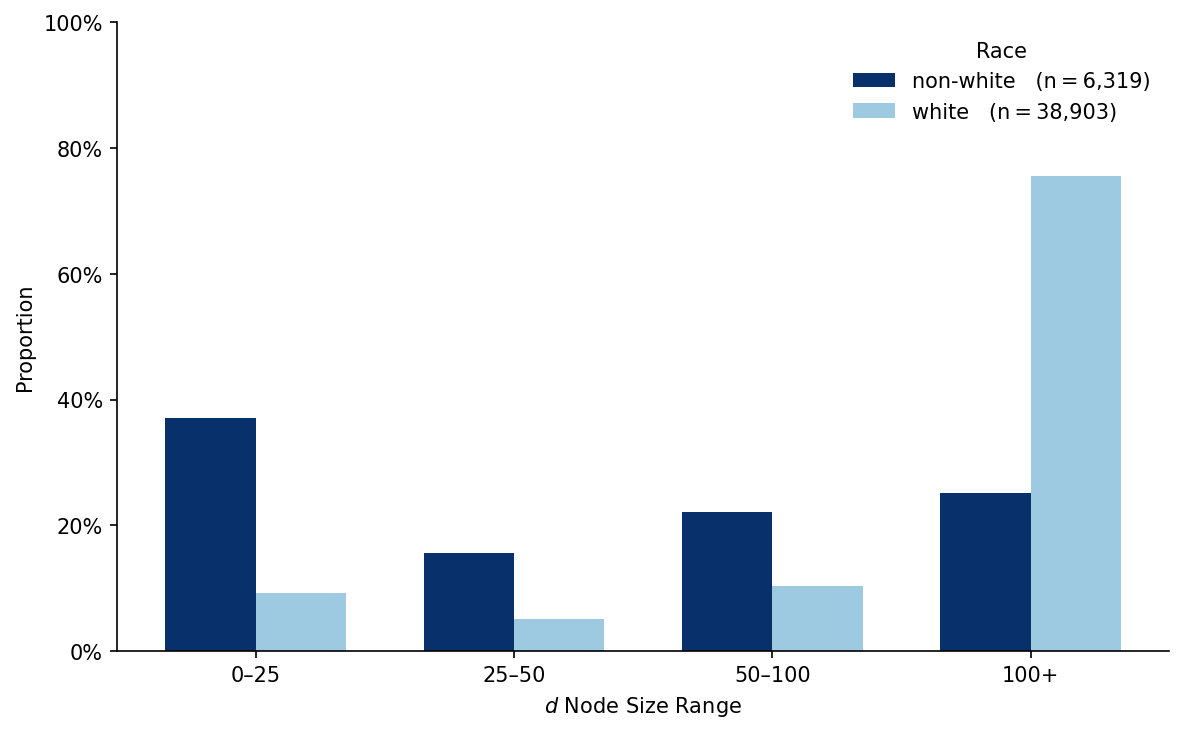}
    \caption{Proportion of individuals in each $d$~node size bin by
    race in the Adult dataset. Non-white individuals are substantially
    more concentrated in the smallest $d$~nodes, where small-sample
    inference effects are most severe.}
    \Description{Grouped bar chart showing the proportion of non-white
    and white individuals in four d-node size bins (0-25, 25-50, 50-100,
    100+) for the Adult dataset. Non-white individuals have a much higher
    proportion in the 0-25 bin (37\% vs 10\%) and a much lower proportion
    in the 100+ bin (25\% vs 76\%).}
    \label{fig:racednodes}
\end{figure}

The same step that improves accuracy, however, can erase fine-grained
distinctions in the data. Minority or intersectional groups are
absorbed into larger aggregate groups, and their distinct behaviour
is overwritten by the majority. The groups most vulnerable to this
are precisely those that are small and sparsely represented, and in
datasets where protected demographic attributes (race, sex, age)
intersect with other features, minority demographics are
disproportionately represented among small $d$~nodes. A larger share
of minority-group $d$~nodes fall into the smallest size bins compared
with majority groups, a pattern that holds across all three of our
benchmark datasets. Figure~\ref{fig:racednodes} illustrates this for
the Adult dataset: 37\% of non-white individuals belong to $d$~nodes
of size 25 or fewer, compared with only 9\% of white individuals,
while 75\% of white individuals belong to nodes of size 100 or more
vs.~25\% of non-white individuals. Regularisation therefore tends to
discard information about the subgroups where prediction quality is
most at risk and where the consequences of poor inference are most
serious.

The issues of small-sample inference therefore remain relevant in 
the era of `big data'. The formal tools required to reason about 
them (Beta posteriors, Beta-Binomial predictive distributions, and 
hypothesis testing for consistency) are developed in 
Section~\ref{sec:preliminaries}.

\subsection{Related Work}

We now position our contribution relative to the three bodies of
literature most relevant to our approach: granular fairness methods,
Bayesian approaches to fairness, and selective prediction with
abstention.

The literature on granular fairness and multicalibration is the 
closest body of work to ours in ambition. Standard fairness metrics 
such as demographic parity and equalised odds consider one 
protected attribute at a time, which can mask substantial 
disparities at the intersection of multiple attributes. Kearns 
et al.\ name this vulnerability `fairness gerrymandering' and 
argue that any serious guarantee must hold over a much more 
granular class of subgroups \cite{kearns2018preventing}.

Multicalibration and multiaccuracy \cite{blum2018multiaccuracy, 
hebert2018} respond to this challenge by requiring that predictive 
performance is reliable across many subgroups simultaneously.
Calibration is the property that predicted probabilities match 
observed frequencies: of items predicted with a 70\% chance of 
being positive, roughly 70\% should actually be positive.
Multicalibration strengthens this by requiring calibration to hold 
for multiple subgroups in the data, while multiaccuracy 
similarly measures accuracy across subgroups, not just on the entire 
dataset. These definitions
are enforced through an iterative auditor-learner loop: an auditor
searches for the subgroup where predictions deviate most from the
target, passes this subgroup to the learner, and the learner adjusts
the model to shrink the gap. Practical implementations include
HappyMap \cite{deng2023happymap} and PMCBoost \cite{lacava2023pmc},
the latter of which we use as a baseline in our experiments.

These methods, however, face three structural limitations. First, 
their guarantees rely on empirical estimates for each subgroup, 
which become noisy when the subgroup is small; corrections based on 
a handful of observations can result in overfitting. Second, because 
auditing every possible feature combination is computationally 
infeasible, practical implementations restrict the search to a 
pre-specified class of subgroups, with the search space growing 
exponentially in the number of protected 
variables~\cite{hebert2018, deng2023happymap, lacava2023pmc}. Large 
portions of the population therefore escape scrutiny. Third, these methods provide no mechanism for 
abstention: the model is required to issue a prediction for every 
input, even when no calibrated prediction can be statistically 
justified. We return to a detailed algorithmic
comparison in Section~\ref{sec:comparison}.

A second relevant body of work applies Bayesian ideas to fairness.
These appear in the literature in two main ways. 
In much of Bayesian ML, priors and posteriors are placed over the 
parameters of a predictive model (for example, the weights of a 
neural network or the coefficients of a regression). The aim is to 
capture epistemic uncertainty: uncertainty about the model itself that 
arises because the training data is
finite and cannot assign a single set of parameters with certainty.
Dimitrakakis et al.\ \cite{dimitrakakis2019bayesian} and Zeng et al.\
\cite{zeng2022bayes} take this parameter-centric view, incorporating
fairness penalties into the Bayesian decision rule. The focus of our
work is different: we use Bayesian inference in its classical sense,
placing posteriors over the population probabilities that generated
the observed data rather than over model parameters. This distinction 
matters because consistency in practice is defined by how the 
model generalises to the population level, not by the stability of 
internal model parameters.

A closely related strand within this literature uses hierarchical
Bayesian models to stabilise subgroup-level estimates. Foulds et al.\
\cite{foulds2020bayesian} assign Beta priors to subgroup target rates in a
Beta-Binomial setting, allowing small groups to draw information from
related groups. Ji et al.\ \cite{ji2023can} extend this to settings
where outcomes are selectively observed (for example, we only learn
whether a borrower repays a loan if the loan is granted), and Perrone
et al.\ \cite{perrone2021fair} extend it to sequential decision-making.
These approaches represent an important step toward accounting for
small-sample uncertainty in fairness assessments, but they remain
frameworks for measuring or guiding fairness rather than complete
classifiers that enforce consistency. They do not provide guarantees across 
every possible subgroup, and none offers a mechanism for abstaining 
when the evidence does not support a prediction.

A further strand uses Bayesian networks to evaluate counterfactual
fairness \cite{chiappa2019path, kusner_loftus_russell_silva_2017}, testing whether predictions
would change if a protected attribute were hypothetically different.
The main weakness of this approach is that it requires a correct and
agreed-upon causal graph, something that is rarely available in
practice and on which domain experts frequently disagree
\cite{chiappa2019causal}.

A third body of work concerns selective prediction and abstention. 
A gap in this literature is the lack of recognition for scenarios 
where the evidence positively rules out every deterministic 
prediction. These are cases where both the all-positive and 
all-negative labels for a subgroup are statistically inconsistent 
with the observed data, as illustrated by the 100-person example 
in the Introduction. Current ML methods, including those explicitly 
designed for fairness, fail to account for these scenarios.

The concept of selective classification, where a model is allowed 
to abstain rather than predict, has a long history. Chow's reject 
option in pattern recognition \cite{chow1970optimum} and later work 
on selective prediction \cite{herbei2006classification, 
elyaniv2010selective, wiener2015agnostic} formalised the trade-off between coverage and 
risk. More recently, Schreuder et al.\ 
\cite{schreuder2021classification} show that certain fairness 
criteria such as equalised odds cannot always be satisfied if the 
classifier is forced to decide on every input. Yin et al.\ 
\cite{yin2024fair} design algorithms that learn abstention regions 
balanced across groups, and Lenders et al.\ 
\cite{lenders2024interpretable} argue that stakeholders need to 
understand not only why a model made a prediction but also why it 
chose not to.

Despite these contributions, existing abstention methods share common
limitations. Most derive the abstention decision from model
confidence scores or user-specified thresholds rather than from a
statistical test of whether any consistent deterministic prediction
exists. Fairness is typically evaluated only for coarse protected
groups, leaving disparities between granular subgroups unaddressed.
And crucially, these methods do not distinguish between abstaining
because the model is uncertain and abstaining because the observed
evidence positively demonstrates that no fair deterministic
prediction is possible.

Confidence-based abstention also has a disparate impact on 
small subgroups. Minorities are, by definition, 
underrepresented, and small samples yield wider posteriors 
and lower model confidence. A confidence threshold therefore 
abstains most often on the smallest subgroups, which tend to 
correspond to the minorities that fairness-aware methods are 
designed to protect. Our framework avoids this. Small $d$~nodes 
with wide posteriors are resolved by consistency with the 
broader subgroups ($v$~nodes) they belong to, not by 
abstention. Abstention is reserved for the cases where the 
evidence itself rules out every deterministic prediction, so 
it highlights the subgroups where no fair prediction is 
possible rather than penalising small ones.

Our framework is distinguished from each of these strands on a
specific point.
It enforces consistency exhaustively across every possible subgroup 
in the data, with no pre-specified subgroup class and no 
computational relaxation of coverage.
The Bayesian inference it applies is classical: posteriors over
the unknown population probability that generated the data, not
over model parameters.
It provides a complete classifier rather than a measurement tool,
and its predictions are guaranteed consistent with the observed
evidence for every subgroup.
Its abstention mechanism is grounded in something different from
model uncertainty: when the data contains enough evidence to
reject both possible deterministic predictions, no prediction is
issued, and this follows directly from the consistency
requirement rather than from any confidence threshold.
These properties are formalised in Section~\ref{sec:framework}.

\section{Bayesian Preliminaries}
\label{sec:preliminaries}

Our approach is a direct application of the classical Bayesian 
framework. We specify a prior over the generating probability of 
the target $T$ within each subgroup of the population, update 
that prior with the observed data to obtain a posterior, and use 
the posterior to reason about predictions. When a single 
prediction must be committed to, the principled Bayesian choice 
is the one that maximises posterior likelihood: the prediction 
most likely to arise in a further sample from the same 
population, given the priors and the observed data. Bayesians 
would generally prefer to return the full posterior rather than 
a point decision; we commit because classification tasks demand 
it, so we select the maximum-posterior option among the 
predictions the data does not rule out.

\subsection{Setup and Notation}
\label{sec:notation}
 
We assume an observed dataset $C$ sampled from some population $U$, where each item is described by categorical values on $m$ attributes. We use $d$ to indicate a complete description of an item in $C$: that is, a specific value for every attribute (e.g.\ [male, office, overtime=yes]). Each such $d$ defines a group of identical data points, which we call a $d$~node. Each $d$~node has an associated count $N_d$ (number of occurrences in $C$), target count $T_d$ (number with $T = 1$), and prediction $P_d$ (the number predicted to have $T = 1$, which can range from $0$ to $N_d$).
 
We can also define subsets of the population by specifying only some attribute values, leaving others unspecified. We call these $v$~nodes (e.g.\ [male, office] with overtime unspecified). A $v$~node comprises multiple $d$~nodes whose members are not necessarily identical. We write $U(v)$ for the subset of $U$ consisting of all $d \in U$ for which $v \subseteq d$.

\subsection{Bayesian Posterior for a Subgroup}
\label{sec:posterior}

For a $d$~node with $N_d$ observations of which $T_d$ have $T = 1$, the unknown population probability $p_d = \Pr(T = 1 \mid d)$ follows the Beta posterior
\begin{equation}
\label{eq:posterior}
p_d \sim \mathrm{Beta}(T_d + a_0,\; N_d - T_d + b_0)
\end{equation}
where $a_0, b_0$ are the prior parameters. Throughout this paper we use the uniform prior $\mathrm{Beta}(1, 1)$, giving
\begin{equation}
\label{eq:posterior_uniform}
p_d \sim \mathrm{Beta}(T_d + 1,\; N_d - T_d + 1).
\end{equation}

The uniform prior reflects our lack of prior knowledge about the target rate and is the standard non-informative choice for inference with binary outcomes. For brevity in what follows we write $a_d = T_d + a_0$ and $b_d = N_d - T_d + b_0$, so the posterior in Equation~\ref{eq:posterior} takes the form $\mathrm{Beta}(a_d, b_d)$.

\subsection{The Beta-Binomial Predictive Distribution}
\label{sec:betabinom}

Given the posterior $p_d \sim \mathrm{Beta}(a_d, b_d)$, the number of target occurrences $K$ in a new sample of size $N$ drawn from the same population follows the Beta-Binomial distribution:
\begin{equation}
\label{eq:betabinom}
K \sim \mathrm{BetaBinomial}(N,\, a_d,\, b_d)
\end{equation}
with probability mass function
\begin{equation}
\label{eq:betabinom_pmf}
\Pr(K = k \mid N, a, b) = \binom{N}{k} \frac{B(k + a,\; N - k + b)}{B(a,\, b)}
\end{equation}
where $B(\cdot,\cdot)$ is the Beta function. This distribution integrates over the uncertainty in $p_d$, giving the predictive distribution for future observations rather than conditioning on a point estimate.

\subsection{Testing Consistency}
\label{sec:consistency}

Our consistency requirement takes the form of a hypothesis test. Each candidate prediction $P_d$ at a $d$~node is treated as a hypothesis about the population target distribution and tested against the observed data under the Beta-Binomial predictive distribution; candidates that are incompatible with the data at significance level $\alpha$ are rejected. Concretely, we ask whether $P_d$ and $T_d$ could plausibly represent target counts in samples drawn from the same population. Given the cumulative distribution function $F_{\mathrm{BB}}$ of $\mathrm{BetaBinomial}(N_d, a_d, b_d)$, a prediction $P_d$ is consistent at significance level $\alpha$ if
\begin{equation}
\label{eq:consistency_test}
\alpha/2 \;\leq\; F_{\mathrm{BB}}(P_d \mid N_d,\, a_d,\, b_d) \;\leq\; 1 - \alpha/2.
\end{equation}

Equivalently, we define for each $d$~node two critical values: $V_{\min}(\alpha, d)$, the smallest integer satisfying $F_{\mathrm{BB}}(V_{\min}) \geq \alpha/2$, and $V_{\max}(\alpha, d)$, the largest integer satisfying $F_{\mathrm{BB}}(V_{\max}) \leq 1 - \alpha/2$. A prediction $P_d$ is consistent if and only if
\begin{equation}
\label{eq:vmin_vmax}
V_{\min}(\alpha, d) \;\leq\; P_d \;\leq\; V_{\max}(\alpha, d).
\end{equation}

These bounds are determined entirely by the observed data and the chosen $\alpha$; they do not depend on any model or learning procedure.

\subsection{Accounting for Heterogeneity}
\label{sec:heterogeneity}

A standard assumption in machine learning is that a dataset 
reflects a single homogeneous population, with observations 
differing only in the values of the recorded variables. For 
large datasets assembled from multiple sources, time periods, 
and collection protocols, this assumption is rarely justified: 
such datasets aggregate observations from materially different 
sub-populations, producing between-sample variation beyond 
what sampling noise alone would predict. This variation is 
known as heterogeneity, and any posterior inference drawn 
from such a dataset must account for it. For example, the COMPAS 
dataset \cite{compas2016} covers two years of recidivism 
assessments in Broward County, during which policing and 
judicial practice may have shifted.

As the sample size in a $d$~node grows, the Beta posterior 
concentrates, eventually implying near-certainty about the 
population probability. This is the correct Bayesian 
conclusion under the homogeneity assumption, but once that 
assumption fails, very large nodes can produce posteriors 
that are overconfident: the narrow interval implied by a 
large $N_d$ understates the true uncertainty about what the 
population probability would be in a new collection.

We address this by placing a lower bound $\tau$ on the variance
of the inferred Beta distribution.
In effect, this caps the effective sample size: for nodes larger
than some threshold, the distribution is scaled so that its
variance does not fall below $\tau$.
For small nodes, the sample-size uncertainty already dominates
and no adjustment is needed.
We fix $\tau = 10^{-5}$ in all experiments, a value large enough
to reflect realistic measurement and temporal drift while
preserving the evidence in medium-sized nodes.
The choice of $\tau$ is discussed further in
Section~\ref{sec:parameters}.

This treatment is one principled way to handle 
heterogeneity, not a definitive solution. Accounting 
for between-sample variation not captured by the data remains 
an open problem in statistical inference, and different 
variance floors, priors, or scaling rules will produce 
different effective sample sizes and prediction boundaries. 
Our choice of $\tau$ therefore reflects a modelling judgement 
that the data alone cannot resolve.

\section{Inference Framework}
\label{sec:framework}

Given the notation and inference tools established in Section~\ref{sec:notation}, we present our framework for statistically consistent prediction. The framework uses two requirements (determinism and statistical consistency) which together determine, for each subgroup in the data, whether a prediction can be made and what that prediction must be.

Although we speak informally of `the prediction' at a given 
$d$~node, the object our framework produces is a joint 
assignment of target labels across every $d$~node in the 
dataset. The $d$~nodes are not treated independently: they are 
coupled through the $v$~node consistency constraints introduced 
in Section~\ref{sec:vnode_consistency}, and the assignment we 
ultimately select is the one most likely to arise, under the 
priors and observed data, in a further sample from the same 
population. Any individual $d$~node prediction should therefore 
be read as a component of this joint assignment rather than as 
a standalone decision.

\subsection{Determinism}
\label{sec:determinism}

A fundamental requirement in any prediction task is determinism: 
identical items must be treated identically. If a decision algorithm 
predicts some value of $T$ for one member of a $d$~node, then 
consistency requires the same prediction for every other member 
of that $d$~node.

The intuitive justification is that when meaningful consequences 
follow from an algorithmic decision (positive from one value of 
$T$, negative from the other), any decision that treats two 
otherwise-identical items differently is inherently unjustifiable. 
The mathematical justification is that when the true label $T$ 
within a $d$~node is generated by a Bernoulli process with 
probability $p$ (whether truly random or driven by unobserved 
differences), the point prediction with the lowest expected 
$0/1$ loss is constant: predict $T = 1$ for all items if 
$p \geq 0.5$, and $T = 0$ otherwise. Any non-deterministic 
assignment has lower predictive accuracy than this constant 
prediction. Determinism therefore limits $P_d$ to either $0$ 
or $N_d$.

\subsection{Statistical Consistency}
\label{sec:prob_consistency}

The second requirement is that a prediction $P_d$ must be
statistically consistent with all available information about $d$
in the sample $C$.
As established in Section~\ref{sec:consistency}, this means
$P_d$ must fall within the $[V_{\min}, V_{\max}]$ interval
derived from the Beta-Binomial posterior at significance level
$\alpha$.

The requirements of determinism and statistical consistency
can conflict.
There are $d$~nodes where the observed target rate $T_d / N_d$
is high, so predicting $P_d = N_d$ (all positive) would be
deterministic and accurate, yet the Beta-Binomial test rules
out $P_d = N_d$ at level $\alpha$ because the sample is large
enough to conclude that a 100\% positive rate is implausible.
The opposite prediction $P_d = 0$ is equally ruled out.
For such nodes no deterministic prediction is consistent with
the evidence, and a consistent algorithm must abstain.
A consistent algorithm at level $\alpha$ therefore assigns
$P_d = N_d$ where that prediction passes the test, $P_d = 0$
where that prediction passes, and abstains otherwise.

\subsection{\texorpdfstring{$d$}{d} Node Categories}
\label{sec:d_categories}

Using the counts and our chosen priors, we estimate $V_{\min}$ and $V_{\max}$ boundaries for each $d$~node (as defined in Section~\ref{sec:consistency}) in order to assign it a category. From these $V_{\min}$ and $V_{\max}$ boundaries, we can categorise each $d$~node based on the prediction that can be made:

\begin{itemize}
    \item $d_0$: $V_{\min} = 0$ and $V_{\max} < N_d$. The only consistent and deterministic prediction is to predict $T = 0$ for all members of the $d$~node.
    
    \item $d_1$: $V_{\min} > 0$ and $V_{\max} = N_d$. The only consistent and deterministic prediction is to predict $T = 1$ for all members of the $d$~node.
    
    \item $d_{am}$: $V_{\min} = 0$ and $V_{\max} = N_d$. Either deterministic prediction is consistent with the data. The node is ambiguous, we must use other information to decide which prediction to make, namely the constraints of the $v$~nodes these $d$~nodes belong to.
    
    \item $d_{nf}$: $V_{\min} > 0$ and $V_{\max} < N_d$. No deterministic prediction that is consistent with the data is possible.
\end{itemize}

\subsection{Abstention}
\label{sec:abstention}

We now address the category of $d$~nodes where no consistent decision
can be made: $d_{nf}$, defined by $V_{\min} > 0$ and $V_{\max} < N_d$.

The abstention mechanism here differs fundamentally from the selective
prediction literature \cite{chow1970optimum, herbei2006classification,
elyaniv2010selective, wiener2015agnostic}, where a model withholds a prediction because its
confidence is low. That form of abstention arises from insufficient
evidence: the posterior is wide and neither deterministic extreme can
be ruled out. Our abstention arises from the opposite situation:
the data contains sufficient evidence to rule out both possible
predictions. The two cases map to distinct node categories in our
framework. A small group with an ambiguous target rate yields a wide
posterior; the consistency interval is broad enough that both
$P_d = 0$ and $P_d = N_d$ fall within it, placing the node in
$d_{am}$ (ambiguous). That node is resolved by $v$~node constraints,
not by abstention. A $d_{nf}$ node is one where the posterior is
narrow enough to reject both extremes.

To see why sample size determines this distinction, consider two nodes
with approximately the same observed target rate of 50\%. In the
first, $N_d = 6$ and $T_d = 3$: the posterior is
$\mathrm{Beta}(4, 4)$ with mean 0.5 and standard deviation 0.18.
The Beta-Binomial consistency interval is wide; at $\alpha = 10^{-5}$
both $P_d = 0$ and $P_d = 6$ remain plausible and the node is
$d_{am}$. In the second, $N_d = 100$ and $T_d = 51$: the posterior
is $\mathrm{Beta}(52, 50)$ with mean 0.51 and standard deviation
0.050. The interval is narrow enough that $P_d = 0$ falls in the
lower tail and $P_d = 100$ falls in the upper tail; both are
rejected at the chosen significance level and the node is $d_{nf}$.
The near-50\% rate alone does not trigger abstention; it is the
combination of a rate near 0.5 and a sample large enough to be
confident about it.

Predicting everyone positive for this $d_{nf}$ node (the strategy
of a decision tree if this pattern appeared in a leaf) would
misclassify roughly half the group; predicting everyone negative
would do the same. Randomised assignment offers no solution, because
all individuals in a $d$~node share identical values across every 
variable. Assigning different predictions to identical individuals
violates the most basic aspect of equal treatment.

In this case, abstention is the only consistent option. By 
refusing to assign an unsupported prediction, the model avoids 
the inconsistent predictions that traditional ML algorithms, 
including most fairness-corrected ones, will issue. In practice, abstaining flags the node for alternative 
action. Stakeholders may route 
the case to a human expert, for instance, a loan underwriter 
or a parole board, or collect richer features that partition 
the node into smaller subgroups where evidence can justify a 
deterministic label. In either scenario abstention operates as 
a safeguard: it prevents the system from issuing demonstrably 
inconsistent predictions while providing an explicit signal 
that additional information or judgement is required.

Our algorithm formally handles these $d_{nf}$ nodes by setting 
their $V_{\min}/V_{\max}$ boundaries to $0$ and $N_d$ 
respectively, treating them the same as $d_{am}$ nodes. This 
results in the $d_{nf}$ nodes being assigned a prediction that 
is consistent with the other constraints as part of the 
predictive process, but when presenting our final output, we 
flag the $d_{nf}$ nodes as `unable to predict'. Therefore, we 
are able to preserve the information provided by the $d_{nf}$ 
nodes, using it to set bounds for their parent $v$~nodes, while 
also highlighting that no consistent prediction exists.

\subsection{Consistency Across \texorpdfstring{$v$}{v} Nodes}
\label{sec:vnode_consistency}

Our consistency requirement extends beyond individual $d$~nodes.
For each $v$~node, the posterior over the probability of $T = 1$
for a randomly sampled member of $U(v)$ is a distribution $f_v$
derived from the corresponding sample and the chosen priors.
This posterior is a weighted combination of the distributions of
the $d$ child nodes of $v$ and cannot be computed in closed form;
we approximate it through sampling as described in
Section~\ref{sec:vnode_distributions}.
An assignment $P$ is globally consistent at level $\alpha$ if,
for every $d$ and $v$~node, the summed predictions fall within
the $[V_{\min}, V_{\max}]$ interval implied by the corresponding
posterior.\footnote{The framework extends straightforwardly to
a multi-class target $T$ by substituting the Dirichlet
distribution for the Beta and following the same reasoning.}

The most stringent consistent assignment $P^*$ corresponds to
the largest $\alpha$ for which the constraint system is
feasible; we write this $\alpha^*$. In practice, we obtain $P^*$
by selecting the largest $\alpha$ at which all constraints remain
satisfied.

Note that the reasoning behind the resulting list of 
predictions is transparent: all evidence is contained in the 
data. For a given individual, we can identify the $d$~node and 
$v$~nodes they belong to and see the evidence behind the 
prediction that was made. This evidence is not based on the 
model's learned associations between the target and variables 
but is obtained directly from the data.

\subsection{Approximating \texorpdfstring{$v$}{v} Node Distributions}
\label{sec:vnode_distributions}

The data points within a $v$~node are not a homogeneous group as they are within a $d$~node, so we cannot directly estimate the Beta-Binomial distribution from the sample as before. Instead, we combine the target distributions of the $d$ child nodes with the distributions of their relative sizes to determine the distribution of the $v$~node. We derive this combination in a generic two-population setting (populations $A$ and $B$, mirroring the case of a $v$~node with two $d$ children) and return to the $v$/$d$ notation when applying the result.

Consider a probability distribution for a parameter $p_{T|A}$, where this parameter represents the probability of some property $T$ occurring in an element drawn randomly from some population $A$. Suppose that prior to drawing a sample from the population $A$ we have a distribution for $p_{T|A}$ of
\begin{equation}
p_{T|A} \sim \mathrm{Beta}(a, b)
\end{equation}
with the density of this Beta distribution at $x$, as before, being
\begin{equation}
f(x \mid a, b) = \frac{1}{B(a,b)} x^{a-1} (1-x)^{b-1}
\end{equation}

If we now see a sample from that population containing $A_N$ elements of which $A_T$ have that target property, then the updated distribution for $p_{T|A}$ that we should infer given that sample (and this prior distribution) is
\begin{equation}
p_{T|A} \sim \mathrm{Beta}(A_T + a,\; A_N - A_T + b)
\end{equation}

Next consider a situation where we have two disjoint populations $A$ and $B$ and where we take $U = A \cup B$ to be the union of those two populations. Assume for simplicity we have the same priors
\begin{align}
p_{T|A} &\sim \mathrm{Beta}(a, b) \\
p_{T|B} &\sim \mathrm{Beta}(a, b)
\end{align}
for $A$ and $B$ and that we have observed samples $A_T / A_N$ and $B_T / B_N$, giving
\begin{align}
p_{T|A} &\sim \mathrm{Beta}(A_T + a,\; A_N - A_T + b) \\
p_{T|B} &\sim \mathrm{Beta}(B_T + a,\; B_N - B_T + b)
\end{align}

Given this, what probability distribution should we infer for the parameter $p_{T|U}$, representing the probability of $T$ occurring in an element drawn randomly from the combined population $U$?

One possibility is to say that, since we have seen a sample of $A_N + B_N$ members of population $U$ of which $A_T + B_T$ have the target property, the distribution we should infer for $p_{T|U}$ is
\begin{equation}
p_{T|U} \sim \mathrm{Beta}(A_T + B_T + a,\; A_N + B_N - A_T - B_T + b)
\end{equation}

This approach contradicts a basic requirement of inference: that the less information we have about something, the more uncertain we must be in our judgments about that thing. Knowing that something was randomly sampled from population $U$ gives us less information than knowing that it was sampled from $A$ (or from $B$): therefore our uncertainty in $p_{T|U}$ must be greater than in $p_{T|A}$ or $p_{T|B}$. The uncertainty associated with a distribution $\mathrm{Beta}(T, N-T)$, however, falls as the values $T$ and $N$ increase (the larger the sample size and target count, the less uncertainty in the inferred probability distribution): the pooled approach therefore gives $p_{T|U}$ a lower uncertainty than $p_{T|A}$ or $p_{T|B}$, contradicting this requirement.

Given a sample from $U$ containing $A_N$ members of $A$ and $B_N$ of $B$ we can consider the parameter $p_{A|U}$, representing the probability of getting a member of $A$ in a random sample from $U$; and since this is a probability it will have the Beta distribution
\begin{equation}
p_{A|U} \sim \mathrm{Beta}(A_N + a',\; B_N + b')
\end{equation}
where $a'$ and $b'$ are our sample size priors. If we knew the values of parameters $p_{T|A}$, $p_{T|B}$ and $p_{A|U}$ we could express $p_{T|U}$ as
\begin{equation}
\label{eq:ptu}
p_{T|U} = p_{T|A}\, p_{A|U} + p_{T|B}\,(1 - p_{A|U})
\end{equation}

Unfortunately we don't know these parameter values: all we have is their density functions
\begin{align}
\Pr(p_{T|A} = x) &= f_{T|A}(x) = f(x;\; A_T + a,\; A_N - A_T + b) \\
\Pr(p_{T|B} = y) &= f_{T|B}(y) = f(y;\; B_T + a,\; B_N - B_T + b) \\
\Pr(p_{A|U} = z) &= f_{A|U}(z) = f(z;\; A_N + a',\; B_N + b')
\end{align}

Our aim, therefore, is to estimate the density function for $p_{T|U}$ (the function that gives us the probability that parameter $p_{T|U}$ has some value $w$).

The expression for $p_{T|U}$ in Equation~\eqref{eq:ptu} is non-linear, and so an analytic expression for the density of $p_{T|U}$ in terms of the densities of its components does not seem to be available. Instead, we can approximate this density numerically by dividing the $0 \ldots 1$ domain into $n$ equal-sized intervals of size $1/n$ and letting $c_i$ be a count associated with each such interval. We then draw triplets of values $x, y, z$ sampled randomly from the density functions for $p_{T|A}$, $p_{T|B}$ and $p_{A|U}$, calculate $w = xz + y(1-z)$ for those values, find the interval $i$ such that $i/n \leq w < (i+1)/n$, and set $c_i = c_i + 1$. If we do this random sampling process $M$ times for some large number $M$, then $c_i / M$ approximates the density of $p_{T|U}$ in the interval $i/n \ldots (i+1)/n$ (the chance that the true value of $p_{T|U}$ will fall into that interval) and we have
\begin{equation}
\Pr(p_{T|U} = w) \approx f_{T|U}(w) = c_i / M \quad \text{where} \quad i/n \leq w < (i+1)/n
\end{equation}

Given a `Beta-like' density function of this form we can approximate the mass function of the analogous Beta-Binomial (giving the probability of getting exactly $k$ successes in a sample of size $N$ where the generating probability has density $f_{T|U}(w)$) as
\begin{equation}
\label{eq:vnode_pmf}
\Pr(k;\, N,\, f_{T|U}) = \sum_{i=0}^{n-1} \mathrm{Bin}(k;\, N,\, p = i/n + 1/2n)\; \frac{c_i}{M}
\end{equation}

Here $p = i/n + 1/2n$ is one of the possible values of the parameter $p_{T|U}$ (approximated by the centre of one of our $n$ intervals) and $c_i/M$ is the probability of that being the true value of the parameter, and $\mathrm{Bin}(k; N, p)$ gives the Binomial probability of getting $k$ successes in a sample of size $N$ when the generating probability is equal to that value $p$.

This gives us an approximation for the Beta distribution of each $v$~node, which can be used to find the Beta-Binomial distribution and $V_{\min}$ and $V_{\max}$ boundaries as before. $v$~nodes with more than two $d$~nodes are handled by combining the distributions of the $d$ child nodes pairwise until we have a single distribution.

In practice, rather than computing every $v$~node distribution directly from its $d$ child nodes (which becomes computationally infeasible for nodes with many ambiguous variables), we utilise the hierarchy of the $v$~nodes. We calculate the distributions of $v$~nodes with a single ambiguous variable (`level one' $v$~nodes) from their $d$ child nodes. For a $v$~node with two ambiguous variables (a `level two' $v$~node) we first determine its $v$ child nodes, i.e.\ `level one' $v$~nodes that are a subset of this parent node. We can then use the distributions of these $v$ child nodes in place of the $d$~nodes to estimate the distribution of the parent $v$~node. We continue this process up the hierarchy until an estimate is found for every $v$~node. Finally, we apply our proxy for heterogeneity to the distribution as necessary and calculate the resulting $V_{\min}$ and $V_{\max}$ boundaries for every $v$ and $d$~node.

\subsection{Constraint Satisfaction}
\label{sec:constraint_sat}

Our list of $v$~nodes now becomes a list of constraints. 
The sum of predictions made for $d$ child nodes of a 
given $v$~node must fall between the $V_{\min}/V_{\max}$ 
boundaries of that $v$~node. A consistent algorithm is one 
whose predictions satisfy all these constraints. We can 
immediately begin by simplifying the entries of this 
list by substituting in values for $d_0$ and $d_1$ nodes 
as they are already decided. This leaves a list of constraints 
which we solve to find predictions for the remaining ambiguous 
$d_{am}$ nodes. Given the number of constraints, a highly 
efficient constraint satisfaction algorithm is required. 
We have chosen to use the Gurobi Optimiser. The software 
takes as input the list of constraints along with an objective 
function. The choice of objective function expresses a preference 
among the feasible assignments; the specific form we use, along 
with a post-solve re-scoring step that refines the choice using 
$v$-node evidence, is described in 
Section~\ref{sec:solution_selection}. 
Consistency with the $d$ and $v$~node constraints is guaranteed 
regardless of which feasible assignment is selected. Gurobi 
reports infeasibility when no feasible assignment exists. Solutions 
may be infeasible for larger $\alpha$ significance levels or for other 
more stringent choices of initial parameters. In these cases, a 
relaxation of $\alpha$ is typically required to find a feasible 
solution.

We combine Gurobi's output with our initial predictions for $d_0$ 
and $d_1$ (and the abstentions for $d_{nf}$) to produce the 
algorithm's final solution. The output is a prediction for each 
observed combination of variables ($d$~node) at which a consistent 
decision is possible at the chosen significance level $\alpha$; 
combinations where no consistent decision exists are flagged. The 
predictions are simultaneously consistent for every sub-combination 
of variables ($v$~node), so any individual the algorithm predicts 
on can trace their prediction directly to the data, and no group 
or subgroup has a prediction statistically inconsistent with the 
observed evidence.

\subsection{Solution Selection}
\label{sec:solution_selection}

The constraints of Section~\ref{sec:constraint_sat} define a
feasible set of solutions but do not determine a unique
assignment. We therefore select in two stages. (i) The Gurobi
mixed-integer program (MIP) maximises a linear $d$~node objective
(Equation~\ref{eq:objective_full}) and returns a pool
$\mathcal{P}$ of the top-ranked feasible solutions. (ii) Each
pool member is then scored by a $v$-node log-likelihood, and
the best-scoring member is the final prediction. The two
objectives are complementary: the $d$-objective is linear and
fits directly into a MIP, while the $v$-node log-likelihood is
nonlinear but can be evaluated cheaply on the
handful of solutions Gurobi returns.

The MIP assigns each $d$~node a weight
\[
w_d \;=\; \log\!\frac{\hat{p}_d}{1 - \hat{p}_d} \;\cdot\; \log(N_d + 1),
\]
where $\hat{p}_d = (T_d + 1)/(N_d + 2)$ is the posterior mean of the
positive-label probability under the uniform $\mathrm{Beta}(1,1)$ prior
\cite{gelman2013bda} (Equation~\ref{eq:posterior_uniform}). The first
factor is the posterior log-odds: positive when $\hat{p}_d > 0.5$ and
negative when $\hat{p}_d < 0.5$, so its sign matches the Bayesian optimal
label at that $d$~node under $0/1$ loss, and its magnitude grows as
$\hat{p}_d$ moves away from $0.5$. Without $v$-node constraints the
MIP would set each free $x_d$ to that Bayesian optimal value; when the
constraints force $x_d$ to take the opposite value the objective is
penalised by $|w_d|$, so the optimiser absorbs these forced deviations on
$d$~nodes whose posterior is closest to $0.5$. The second factor,
$\log(N_d + 1)$, scales each weight by $d$~node size, so larger
$d$~nodes carry more influence in the objective but not linearly more.

For each $v$~node $v$ with Beta-Binomial predictive PMF $M_v$
(Section~\ref{sec:vnode_distributions}) and $d$-children
$\mathrm{dc}(v)$, let
$t_v(\mathbf{x}) = \sum_{d \in \mathrm{dc}(v)} T_d \, x_d$
denote the count of observed positives inside $v$ that assignment
$\mathbf{x}$ labels positive. The final prediction is
\begin{equation}
\label{eq:selector}
\mathbf{x}^{\star} \;=\; \arg\max_{\mathbf{x} \in \mathcal{P}}
\;\sum_{v \in \mathcal{V}} \log M_v\!\big(t_v(\mathbf{x})\big).
\end{equation}
Writing
$\ell_v(\mathbf{x}) = \sum_{v \in \mathcal{V}} \log M_v(t_v(\mathbf{x}))$
and $L_v(\mathbf{x}) = \exp(\ell_v(\mathbf{x}))$, differences in
$\ell_v$ correspond to likelihood ratios under this $v$-node scoring
model:
\[
\frac{L_v(\mathbf{x}_i)}{L_v(\mathbf{x}_j)}
\;=\;
\exp\!\left(\ell_v(\mathbf{x}_i) - \ell_v(\mathbf{x}_j)\right).
\]
Small gaps in $\ell_v$ therefore correspond to large likelihood ratios
between competing solutions \cite{gelman2013bda,kassraftery1995}.

Because $\mathbf{x}^{\star}$ is drawn from $\mathcal{P}$, it satisfies
every $d$~node and $v$~node $V_{\min}/V_{\max}$ bound by construction.
Full details (pool configuration, treatment of $d_{nf}$ nodes, and
numerical safeguards) are in Appendix~\ref{app:pseudocode}.

\section{Algorithm}
\label{sec:algorithm}
 
We now describe the implementation of our inference framework as a practical algorithm. Table~\ref{tab:node_counts} quantifies the scale of the problem for our benchmark datasets. Given that most current fair ML approaches rarely cover more than a handful of groups, our framework offers a substantially more granular enforcement of consistency.
 
\begin{table}[ht]
\centering
\begin{tabular}{lccc}
\hline
Dataset & No.\ of Entries & No.\ of $d$~Nodes & No.\ of $v$~Nodes \\
\hline
Adult & 45,222 & 1,376 & 81,380 \\
COMPAS & 5,388 & 1,485 & 32,483 \\
Bank Marketing & 45,211 & 4,432 & 142,030 \\
\hline
\end{tabular}
\caption{Number of total entries, $d$ and $v$~nodes for each dataset.}
\label{tab:node_counts}
\end{table}
 
Our code is written in Python 3.9.6 using pandas 2.3.3, NumPy 2.0.2, SciPy 1.13.1, scikit-learn 1.6.1 and Gurobi Optimizer 12.0.3 (via \texttt{gurobipy}), on macOS 26.3. The model finds a solution for each of our benchmark datasets with modest computational resources (2020 MacBook Pro M1, 16GB RAM). The full implementation, together with scripts to download the benchmark datasets from their public sources and regenerate every table and figure in this paper, is available at \url{https://github.com/owenon7/fairbayesian} under an MIT licence.

\subsection{Parameters}
\label{sec:parameters}
 
The algorithm has four user-set parameters; none are tuned on a hold-out set or selected to maximise accuracy.

\begin{description}
\item[Significance level $\alpha$.] Controls the stringency of the consistency requirements. If $\alpha$ is too high, the $V_{\min}/V_{\max}$ intervals contract and the constraint system becomes infeasible; if too low, the intervals widen until almost any prediction satisfies them. We select the largest $\alpha$ for which the constraints remain jointly satisfiable, starting from a conservative default and relaxing until feasibility is achieved.

\item[Prior $\mathrm{Beta}(a_0, b_0)$.] Defaults to the uniform $\mathrm{Beta}(1,1)$; domain knowledge may justify alternatives, but such choices should be explicitly motivated.

\item[Heterogeneity floor $\tau$.] Caps the effective sample size for large nodes (see Section~\ref{sec:heterogeneity}); we fix $\tau = 10^{-5}$ in all experiments.

\item[Density resolution.] Controls the number of grid points used for numerical density approximations; we use 1000 intervals throughout.
\end{description}

\subsection{Pipeline}
\label{sec:pipeline}
 
The algorithm proceeds in four stages. Detailed pseudocode is provided in Appendix~\ref{app:pseudocode}.
 
\textbf{Stage 1: $d$~node classification.}
The data is summarised into $d$~nodes. For each, we compute the Beta posterior (Equation~\ref{eq:posterior_uniform}), derive the Beta-Binomial predictive distribution (Equation~\ref{eq:betabinom_pmf}), and determine the $V_{\min}/V_{\max}$ bounds (Equation~\ref{eq:vmin_vmax}). Each $d$~node is then assigned to one of the four categories defined in Section~\ref{sec:d_categories}: $d_0$, $d_1$, $d_{am}$, or $d_{nf}$.
 
\textbf{Stage 2: $v$~node generation and distribution estimation.}
We generate the set of $v$~nodes present in the data and reduce it by removing redundant constraints (nodes with a single $d$ child, duplicate child sets, or trivially satisfied bounds). For each remaining $v$~node, we approximate its Beta distribution by combining the distributions of its child nodes via sampling, as described in Section~\ref{sec:vnode_distributions}, proceeding hierarchically from level-one $v$~nodes upward.
 
\textbf{Stage 3: Constraint satisfaction.}
The $V_{\min}/V_{\max}$ bounds for all $d$ and $v$~nodes are formatted as constraints over binary decision variables (one per $d$~node). We substitute the known values for $d_0$ and $d_1$ nodes and solve using the Gurobi Optimiser with the objective function in Equation~\ref{eq:objective_full}; the post-solve selection of a final assignment from the solver's pool of feasible solutions is described in Section~\ref{sec:solution_selection}. One edge case requires attention before solving: $v$~nodes dominated by $d_{nf}$ child nodes (those where more than 50\% of member individuals belong to $d_{nf}$ nodes) can produce infeasible constraints. For these nodes we widen the $V_{\min}/V_{\max}$ boundaries by the aggregate size of their $d_{nf}$ children, preventing the abstaining nodes from over-constraining the system while still using their evidence. If no feasible solution exists after this adjustment, $\alpha$ is relaxed and the process restarts.
 
\textbf{Stage 4: Output.}
The solution assigns a prediction to each $d$~node. Nodes classified as $d_{nf}$ are flagged as unable to predict. The result is a complete, transparent assignment: for any individual, the $d$ and $v$~nodes they belong to and the evidence behind the prediction can be directly inspected.

\subsection{Comparison to Existing Approaches}
\label{sec:comparison}

Our consistency definition can be viewed as a more stringent
form of multicalibration. We replace the empirical calibration
metric with a Bayesian-posterior check against the observed
sample, and we apply this check to every possible subgroup in
the dataset. Multicalibration, by contrast, is limited by
computational constraints to a pre-specified subset of protected
groups.
We now describe how this difference manifests
in practice.

The central computational challenge is that the
number of intersectional subgroups grows
exponentially with each additional protected
attribute~\cite{kearns2018preventing}.
In the standard multicalibration framework,
as implemented by HappyMap~\cite{deng2023happymap}
and PMCBoost~\cite{lacava2023pmc}, an auditor-learner
loop identifies the subgroup whose calibration is
most violated and adjusts the predictor to correct
it, repeating until all subgroups are within
tolerance. H\'{e}bert-Johnson
et al.~\cite{hebert2018} show this requires
$O(|\mathcal{C}|/(\alpha^3\lambda^2\gamma))$
iterations, where $|\mathcal{C}|$ is the number
of candidate subgroups, $\alpha$ the calibration
tolerance, $\lambda$ the bin width, and $\gamma$
the minimum group
probability~\cite{lacava2023pmc}.
Because $|\mathcal{C}|$ grows exponentially with
each new protected attribute, so does the required
number of iterations.
Each iteration also requires finding the single
worst-violating subgroup across all of
$\mathcal{C}$, a step the authors of HappyMap
note they do not address directly, observing that
the search is ``at least as hard as agnostic
learning''~\cite{deng2023happymap}.
Both the number of iterations and the cost of
each iteration therefore grow with the protected
attribute set.

To remain computationally tractable, existing
methods work with a fixed, pre-enumerated
candidate class $\mathcal{C}$ and a relaxed
calibration threshold, treating any subgroup
below tolerance as satisfactorily calibrated.
This is a well-motivated design choice under
the given computational constraints, but it
means that guarantees apply only to subgroups
that were pre-specified or triggered the auditor.

Our model takes a different approach.
It enforces consistency across all groups in the
data (abstaining where this is not possible)
without pre-restricting the subgroup class or
relaxing the consistency requirement.
The difference in scope is substantial in
practice: PMCBoost, the state-of-the-art method
we use as a baseline, experiments with at most
three attributes (race/ethnicity, gender, and
insurance type), yielding approximately 14--28
subgroup intersections~\cite{lacava2023pmc}.
Even at this limited scale, satisfying proportional
multicalibration guarantees via the standard
MCBoost algorithm requires up to $1{,}000\times$
more iterations for typical parameter settings
($\rho = 0.1$), and empirically around five times
as many updates to reach comparable calibration
performance~\cite{lacava2023pmc}.

\subsection{Unseen Data}
\label{sec:unseen}

Our framework is developed for the setting where predictions
are derived from observed data, and this is the primary
focus of the paper.
For completeness, we note that a preliminary mechanism exists
for handling test instances belonging to $d$~nodes not
present in the training set: the consistency constraints
are re-evaluated after tentatively assigning each candidate
label, and the outcome that preserves feasibility is
selected (or the instance is abstained from if neither
or both do).
A full description of this procedure is given in
Appendix~\ref{app:unseen}.
Extending this mechanism to a fully general and efficient
deployment setting is left as future work.

\section{Experiments}
\label{sec:experiments}

We evaluate our model using three public benchmark datasets. Because our approach is designed from first principles to satisfy a statistical consistency notion and to abstain whenever that goal cannot be met, familiar accuracy-focused metrics are less appropriate as primary measures. Instead, our evaluations centre on our notion of statistical consistency across all $d$ and $v$~nodes.

Alongside our algorithm we evaluate three baselines: a decision tree (DT), a neural network (NN), and Proportional Multicalibration (PMC) applied to a logistic regression model. For each dataset we cover three aspects: (1) node-level consistency analysis, (2) illustrative case studies, and (3) accuracy, multiaccuracy and multicalibration as sanity checks.

Our primary evaluation asks what each model has learned from the
data: specifically, whether its predictions are statistically
consistent with the observed evidence across all $d$ and $v$
nodes.
This question requires every model to predict on the full dataset
so that predictions can be compared against the complete node
structure.
A held-out test set would fragment $d$~nodes, the very units
against which consistency is measured. Because each $d$~node is
defined by an exact combination of feature values, a random
row-wise split scatters members of the same $d$~node across
train and test, redefining the units of evaluation in the
process. The $V_{\min}/V_{\max}$ bounds that constitute the
consistency criterion are properties of $d$~nodes as observed
in the full dataset; reducing the data on which they are
computed reduces the precision of the evaluation itself, not
just the model.

The key difference between the models lies in how they use the
data before prediction.
Our model receives the entire dataset directly.
It does not require a train/test split or any form of
regularisation because its protection against overfitting is
structural: the Bayesian posterior widens for small samples,
producing wider $V_{\min}/V_{\max}$ bounds that lead to
abstention or ambiguity rather than confident but unsupported
predictions.
Presenting it with more data makes the posteriors more precise,
which is the desired behaviour.
The baseline models (DT, NN, and PMC) follow the standard ML
pipeline: hyperparameters are selected via ten-fold stratified
cross-validation on an 80/20 training split, and the best
configuration is retrained on the full training set.
These models are then used to predict on all entries.
Roughly 80\% of each baseline's predictions are therefore
in-sample, where its accuracy and consistency should be at
their most favourable. Despite this advantage, the results
below show that the baselines still produce predictions that
are statistically inconsistent with the observed evidence on
a substantial proportion of subgroups. These inconsistencies
cannot be attributed to out-of-sample generalisation error:
they appear on data the baselines were directly fitted to.

We compare results via the $d$ and $v$~node structure.
The deeper argument that our framework cannot overfit in the
conventional sense even when given the full dataset is
developed in Section~\ref{sec:overfitting}.

\subsection{Datasets}
\label{sec:datasets}

We evaluate on three public benchmark datasets central to the algorithmic fairness literature: Adult~\cite{adultkohavi} (income prediction), COMPAS~\cite{compas2016} (recidivism risk), and Bank Marketing~\cite{zafar2017fairness1, zafar2019fairness} (product uptake). Together they span employment, criminal-justice, and consumer-finance settings, each with documented evidence of disparate treatment across sensitive attributes. All three contain exclusively categorical or easily discretised features, allowing straightforward creation of $d$ and $v$~nodes. We apply the same cleaning pipeline to every dataset: removal of incomplete or duplicated records, binning of continuous variables using thresholds reported in earlier fairness studies, and one-hot encoding for the baseline ML models.

The protected attributes are left in as variables, both to keep the setting realistic (removing them rarely prevents proxy discrimination) and to allow explicit inspection of how models make use of sensitive information. The protected categories are: sex and race for Adult; sex and race for COMPAS; and marital status and age for Bank Marketing.

\begin{table}[ht]
\centering
\small
\begin{tabular}{l|ll}
\textbf{Attribute} & \textbf{Description} & \textbf{Unique Values} \\
\hline
target & Income level & \{0, 1\} \\
sex & Gender & \{Male, Female\} \\
age\_cat & Categorised age & \{25-60, <25, >60\} \\
capital\_gain\_cat & Categorised capital gains & \{$\leq$5000, >5000\} \\
hours\_per\_week\_cat & Categorised work hours & \{40-60, <40, >60\} \\
workclass\_cat & Employment type & \{non-private, private\} \\
education\_cat & Education level & \{high, low\} \\
marital\_status\_cat & Marital status & \{other, married\} \\
native\_country\_cat & Country of origin & \{US, non-US\} \\
race\_cat & Racial category & \{white, non-white\} \\
occupation\_cat & Occupation category & \{office, heavy-work, other\} \\
\end{tabular}
\caption{Adult dataset variables and categories.}
\label{tab:adultcat}
\end{table}

\begin{table}[ht]
\centering
\small
\begin{tabular}{l|ll}
\textbf{Variable} & \textbf{Description} & \textbf{Categories} \\
\hline
target & Recidivism outcome & \{0, 1\} \\
sex & Gender & \{Male, Female\} \\
race\_cat & Racial category & \{non-caucasian, caucasian\} \\
age\_cat & Age group & \{>45, 25--45, <25\} \\
c\_charge\_desc & Primary charge & 83 different crimes \\
c\_charge\_degree & Charge degree & \{F, M\} \\
priors & No.\ of prior offences & \{0, 1--5, >5\} \\
juv\_fel & Juvenile felonies & \{0, 1--5, >5\} \\
juv\_misd & Juvenile misdemeanours & \{0, 1--5, >5\} \\
\end{tabular}
\caption{COMPAS dataset variables and categories.}
\label{tab:compascat}
\end{table}

\begin{table}[ht]
\centering
\small
\begin{tabular}{l|ll}
\textbf{Attribute} & \textbf{Description} & \textbf{Unique Values} \\
\hline
target & Customer response & \{0, 1\} \\
age\_cat & Categorised age & \{25-60, <25 or >60\} \\
marital & Marital status & \{married, single, divorced\} \\
education & Education level & \{tertiary, secondary, unknown, primary\} \\
housing & Housing loan & \{yes, no\} \\
contact & Communication type & \{unknown, cellular, telephone\} \\
month & Month of last contact & \{jan, feb, mar, \ldots\} \\
duration\_cat & Categorised call duration & \{121-600, $\leq$120, >600\} \\
pdays\_cat & Days since last contact & \{$\leq$30, 31-180, >180\} \\
poutcome & Previous campaign outcome & \{unknown, failure, other, success\} \\
\end{tabular}
\caption{Bank Marketing dataset variables and categories.}
\label{tab:bmvars}
\end{table}

\subsection{Baseline Models}
\label{sec:baselines}

We compare our model with three reference approaches:
a decision tree (DT), a neural network (NN), and Proportional Multicalibration (PMC), a post-processor
applied to logistic regression.
These baselines represent widely used predictive and
fairness-adjusted pipelines, and illustrate how each
behaves when assessed through our statistical
consistency criteria.
All implementations use scikit-learn~\cite{sklearn}.

Each baseline is trained on the same 80/20 stratified
split of every dataset (\textit{random\_state}~=~42).
Categorical variables are one-hot encoded.
Hyperparameters are selected via ten-fold stratified
cross-validation on the training set; the best
configuration is then re-fitted on the full training
set and evaluated on the held-out test set.\footnote{%
For our model no cross-validation is required;
it is evaluated on the complete dataset.}
The full search grids and final selected configurations
for each dataset are reported in
Appendix~\ref{app:hyperparams}.

\textbf{Decision Tree.}
We use scikit-learn's \texttt{DecisionTreeClassifier}.
The search grid covers impurity criteria
\{gini, entropy\}, maximum depths
\{None, 5, 10, 15\}, minimum samples per split
\{2, 5, 10\}, minimum samples per leaf
\{1, 2, 5\}, maximum features
\{None, sqrt, log2\}, and cost-complexity
pruning penalties \{0.0, 0.001, 0.01, 0.1\}.
Selection uses 10-fold stratified CV scored
by accuracy (\textit{random\_state}~=~42).

\textbf{Neural Network.}
We use scikit-learn's \texttt{MLPClassifier}.
The search grid covers hidden-layer configurations
\{$(10,10)$, $(100,10)$, $(50,50)$\},
activation functions \{relu, logistic\},
$\ell_2$ penalties \{$10^{-4}$, $10^{-3}$\},
solver \{adam\}, and learning rate schedule
\{adaptive\}, with a maximum of 200
training iterations.

\textbf{Proportional Multicalibration (PMC).}
Following~\cite{lacava2023pmc}, we apply the
\texttt{MultiCalibrator} as a post-processing step
to a logistic regression base learner
(\texttt{LogisticRegression}, solver \texttt{lbfgs},
\textit{max\_iter}~=~500).
Predicted probabilities are thresholded at the
dataset base rate to produce binary predictions.
The auditor is initialised with the protected
attributes for each dataset; $n\_\text{bins} = 5$;
all other parameters use the library defaults.

\subsection{Evaluation Metrics}
\label{sec:eval}

\begin{itemize}
    \item \textbf{$d_{0/1}$ Node Consistency Error}: The proportion of a model's predictions that violate the unique consistent label for $d_0$ and $d_1$ nodes. By construction our model incurs zero error; any positive value for a baseline indicates it contradicts the clearest evidence in the data.
    
    \item \textbf{$d_{nf}$ Node Consistency Error}: The fraction of the dataset contained in $d_{nf}$ nodes, where neither deterministic label is consistent. As none of the baselines can abstain, this also represents the proportion receiving necessarily inconsistent predictions. Our model abstains on every $d_{nf}$ node.
    
    \item \textbf{$v$~node Consistency Error}: The proportion of $v$~nodes whose aggregated prediction falls outside the $V_{\min}/V_{\max}$ bounds. This reveals intermediate-level contradictions not visible at the $d$~node level.
\end{itemize}

\subsection{Alpha Sensitivity}
\label{sec:alpha}

The significance level $\alpha$ governs the width of the $[V_{\min}, V_{\max}]$ consistency intervals: larger values contract them, imposing stricter requirements on each subgroup; smaller values widen them, relaxing requirements until almost any prediction is admitted. As $\alpha$ decreases from a high value, a feasibility threshold $\alpha^*$ is crossed at which a globally consistent solution first becomes possible; below that threshold, the system remains feasible but with progressively fewer active constraints.

Each active constraint encodes a statistical requirement for a specific subgroup: that the aggregate prediction for its members falls within the subgroup's consistency bounds. A constraint that becomes trivially satisfied at lower $\alpha$ (because the widened interval spans the full prediction range) is no longer placing any effective restriction on predictions, meaning the corresponding subgroup is excluded from statistical enforcement. Selecting $\alpha = \alpha^*$ therefore retains the maximum number of constraints that can be jointly satisfied, enforcing as much subgroup-level statistical information as possible while still guaranteeing a feasible solution.

Table~\ref{tab:alpha_sweep} illustrates this on the Adult dataset. For $\alpha \geq 5 \times 10^{-7}$ the system is infeasible: the consistency intervals are too narrow for any prediction assignment to satisfy all subgroup bounds simultaneously. Feasibility is first achieved at $\alpha^* = 10^{-7}$, with 56,833 active constraints. Relaxing $\alpha$ by two orders of magnitude to $10^{-9}$ reduces this to 52,334 constraints, a reduction of around 8\% that corresponds to the loss of statistical enforcement across several thousand subgroup bounds. The corresponding values on COMPAS and Bank Marketing are $\alpha^* = 2 \times 10^{-6}$, with 11,291 and 70,001 active constraints respectively. At $\alpha^*$, the $v$~node log-likelihood selector (Section~\ref{sec:solution_selection}) returns a unique final prediction from the Gurobi solution pool. To check that this is not a near-tie among essentially equivalent feasible assignments, we also compare the best and second-best pool members by $v$~node log-likelihood. Table~\ref{tab:vll_gap} shows that the selected solution is decisively separated from the runner-up on all three datasets.

\begin{table}[ht]
\centering
\small
\begin{tabular}{r|ccc}
$\alpha$ & \textbf{Feasible} & \textbf{Active Constraints} & \textbf{Unique (N@Best)} \\
\hline
$10^{-3}$               & No  & ---    & --- \\
$10^{-4}$               & No  & ---    & --- \\
$10^{-5}$               & No  & ---    & --- \\
$10^{-6}$               & No  & ---    & --- \\
$5 \times 10^{-7}$      & No  & ---    & --- \\
$10^{-7}$               & Yes & 56,833 & 1   \\
$10^{-8}$               & Yes & 54,305 & 1   \\
$10^{-9}$               & Yes & 52,334 & 1   \\
\end{tabular}
\caption{Alpha sensitivity analysis on the Adult dataset. Active constraints is the number of $v$~node constraints after pruning trivially satisfied bounds. N@Best = 1 indicates the final prediction is unique at every feasible $\alpha$ after the $v$~node log-likelihood selector is applied (Section~\ref{sec:solution_selection}).}
\label{tab:alpha_sweep}
\end{table}

\begin{table}[ht]
\centering
\small
\begin{tabular}{l|rr}
\textbf{Dataset} & $\Delta \ell_v(1,2)$ & \textbf{Likelihood Ratio} \\
\hline
Adult & 36.16 & $5.08 \times 10^{15}$ \\
COMPAS & 12.65 & $3.12 \times 10^{5}$ \\
Bank Marketing & 35.41 & $2.40 \times 10^{15}$ \\
\end{tabular}
\caption{Separation between the best and second-best feasible
solutions at $\alpha^*$, ranked by $v$~node log-likelihood, using a
diagnostic pool of the top 1000 Gurobi solutions by objective score. The likelihood ratio is
$\exp(\Delta \ell_v(1,2))$, comparing the selected solution with the
runner-up.}
\label{tab:vll_gap}
\end{table}

On Adult, for example, the best solution has
$\ell_v=-461{,}627.12$ and the runner-up has
$\ell_v=-461{,}663.28$, a gap of 36.16. This corresponds to a
likelihood ratio of approximately $5 \times 10^{15}$ in favour of the
selected solution. The gap arises because a small change at the
$d$~node level propagates through the dense hierarchy of overlapping
$v$~nodes: the top two Adult solutions differ on only two $d$~node
predictions, but those changes alter the implied target counts in many
$v$~nodes. Thus the selected solution is not an arbitrary choice; 
the $v$~node likelihood makes a clear distinction
between the best and second-best solutions.

This motivates our selection rule: use the largest $\alpha$ 
for which the system is feasible, found via binary search. 
This is not a hyperparameter choice in the conventional ML 
sense; $\alpha^*$ is determined entirely by the data and the 
constraint structure, and tightening it further would render 
the system inconsistent rather than improving prediction quality.

From a fully Bayesian perspective, one might reasonably ask why we 
do not pick $\alpha$ to maximise the $v$~node log-likelihood $\ell_v$ 
of the selected solution, rather than the largest feasible value. The 
answer is that $\ell_v$ values at different $\alpha$ are not 
directly comparable: each $\alpha$ corresponds to a different 
feasible set with different constraints, so the likelihoods 
are evaluated on effectively different models of the data. 
Empirically, the direction of $\ell_v$ with $\alpha$ is not 
even consistent across our three benchmarks: as $\alpha$ 
decreases below $\alpha^*$, $\ell_v$ trends downward on Adult, 
is non-monotonic on COMPAS, and trends upward on Bank Marketing. 
Three datasets yielding three different directions is itself 
evidence that these likelihoods are not meaningfully optimisable 
over $\alpha$. The principled choice is therefore the largest 
feasible $\alpha$: the strongest evidence threshold at which a 
consistent assignment exists.

\subsection{Results: Node-Level Consistency}
\label{sec:node_results}

\subsubsection{\texorpdfstring{$d_{0/1}$}{d01}~node Consistency}

Table~\ref{tab:d01Error} reports the proportion of individuals whose assigned prediction violates the $d_0$ or $d_1$ requirement. Our model achieves zero error on every dataset by design. The DT and NN achieve low error overall, but still produce some predictions inconsistent with clear statistical evidence, particularly on COMPAS (up to 1.3\%). The PMC model shows the largest deviations, especially on Bank Marketing (17.0\%), because its multiplicative calibration step can inflate or deflate probabilities to satisfy calibration targets that conflict with $d$~node-level evidence.

\begin{table}[ht]
\centering
\small
\begin{tabular}{l|rrr}
\textbf{Model} & \textbf{Adult} & \textbf{COMPAS} & \textbf{Bank} \\
\hline
FB  & 0.00\% & 0.00\% & 0.00\% \\
DT  & 0.00\% & 0.58\% & 0.37\% \\
NN  & 0.00\% & 1.28\% & 0.19\% \\
PMC & 0.68\% & 1.28\% & 16.97\% \\
\end{tabular}
\caption{$d_{0/1}$ node consistency error (percentage of instances).}
\label{tab:d01Error}
\end{table}

\subsubsection{\texorpdfstring{$d_{nf}$}{dnf}~node Prevalence and Abstention}

Table~\ref{tab:dnf_summary} reports the share of each dataset falling inside $d_{nf}$ nodes. Any deterministic prediction on these nodes is necessarily inconsistent with the evidence. Our model abstains on every such instance; the baselines always predict.

\begin{table}[ht]
\centering
\small
\begin{tabular}{l|rrr}
\textbf{} & \textbf{Adult} & \textbf{COMPAS} & \textbf{Bank} \\
\hline
Total instances & 45,222 & 5,388 & 45,211 \\
In $d_{nf}$ nodes & 22,798 & 516 & 5,114 \\
$d_{nf}$ proportion & 50.4\% & 9.6\% & 11.3\% \\
No.\ of $d_{nf}$ nodes & 63 & 7 & 32 \\
\end{tabular}
\caption{Prevalence of $d_{nf}$ nodes across datasets.}
\label{tab:dnf_summary}
\end{table}

The Adult dataset has a notably high $d_{nf}$ proportion (50.4\%), meaning that a consistent prediction is impossible for half the individuals. A $d_{nf}$ proportion this high calls into question whether the available variables are sufficient for reliable inference in this domain.

\subsubsection{\texorpdfstring{$v$}{v}~node Consistency}

Table~\ref{tab:vnode_error} reports the proportion of $v$~nodes whose aggregated prediction falls outside the $V_{\min}/V_{\max}$ bounds. Our model satisfies all $v$~node constraints by construction.

\begin{table}[ht]
\centering
\small
\begin{tabular}{l|rcccc}
\textbf{Dataset} & \textbf{Active $v$~Nodes} & \textbf{FB} & \textbf{DT} & \textbf{NN} & \textbf{PMC} \\
\hline
Adult & 53,357 & 0.00\% & 1.64\% & 2.37\% & 2.71\% \\
COMPAS & 6,247 & 0.00\% & 14.44\% & 7.33\% & 7.73\% \\
Bank Marketing & 56,169 & 0.00\% & 5.83\% & 1.73\% & 43.46\% \\
\end{tabular}
\caption{Percentage of $v$~nodes with inconsistent aggregated predictions.}
\label{tab:vnode_error}
\end{table}

All three baselines produce predictions that are statistically inconsistent for a significant proportion of $v$~nodes. The DT on COMPAS shows inconsistency in over 14\% of active $v$~nodes, and PMC on Bank Marketing shows an especially striking 43\%, despite the latter being explicitly optimised for a fairness objective.

\subsection{Illustrative Examples}
\label{sec:examples}

We present three illustrative examples that demonstrate the types of inconsistencies produced by baseline models.

\subsubsection{$d_{0/1}$ Violation: Bank Marketing}

The $d$~node \texttt{\{age\_cat = 25--60; marital = single; education = secondary; housing = yes; contact = unknown; month = May; duration\_cat = 121--600; pdays\_cat $\leq$ 30; poutcome = unknown\}} contains 737 individuals, of whom only 5 subscribed to a term deposit ($5/737$). The $V_{\min}/V_{\max}$ interval is $[0, 34]$, classifying this as $d_0$: the only statistically consistent assignment is the non-target label. Nevertheless, the PMC predictor marks all 737 cases as positive, dramatically overshooting the empirical base rate and $V_{\min}/V_{\max}$ interval, likely a consequence of its multiplicative update step overcompensating due to underprediction in overlapping slices elsewhere.

\subsubsection{$d_{nf}$ Abstention: COMPAS}

Consider the $d_{nf}$ node \texttt{\{Male, non-caucasian, 25--45, Battery,}  \newline
\texttt{Misdemeanour, priors = 1--5, juv\_fel = 0, juv\_misd = 0\}}, which contains 126 defendants, 45\% of whom reoffended within two years. Despite this, all three baselines predict non-target for every individual in the node. There is sufficient evidence to say that either deterministic prediction is inconsistent with the available information. The only consistent option is to abstain. Roughly 10\% of the data in COMPAS belongs to a $d_{nf}$ node, meaning that 1 in 10 defendants will receive a necessarily inconsistent prediction from any model that does not abstain.

In the criminal justice context, inconsistent target predictions unfairly harm defendants, while inconsistent non-target predictions unfairly impact victims.

\subsubsection{\texorpdfstring{$v$}{v}~node Inconsistency: Adult}

Consider the $v$~node \texttt{\{sex = Female, capital\_gain $\leq$ 5000, education = high, marital = married, race = non-white, age\_cat = 25--60, hours\_per\_week = 40--60, occupation = office\}} with the remaining attributes ambiguous. This node contains 103 individuals, 34 of whom have the target. The $V_{\min}/V_{\max}$ interval is $[2, 67]$.

\begin{table}[ht]
\centering
\small
\begin{tabular}{cccccccc}
\textbf{Count} & \textbf{Target} & $V_{\min}$ & $V_{\max}$ & \textbf{FB} & \textbf{DT} & \textbf{NN} & \textbf{PMC} \\
\hline
103 & 34 & 2 & 67 & 10 & 0 & 0 & 73 \\
\end{tabular}
\caption{Summary of an illustrative $v$~node on Adult: non-white married women, 25--60, in office roles.}
\label{tab:FM2560}
\end{table}

Both the DT and NN predict zero target outcomes, falling below $V_{\min} = 2$ and violating the consistency requirement from below, while PMC predicts 73 positives, overshooting $V_{\max} = 67$ from above. Our model predicts 10 positives, squarely inside $[V_{\min}, V_{\max}]$. The DT and NN likely assign non-target because regularisation prevents them from learning the finer subgroup structure; our model predicts target for several of the constituent $d$~nodes specifically to maintain consistency with this and other overlapping $v$~node constraints. This systematic underprediction disadvantages an already marginalised subgroup (non-white married women in office occupations).

\subsection{Performance Comparison}
\label{sec:performance}

The consistency analyses above establish how each method behaves with respect to node-level requirements. We now report accuracy, multiaccuracy, and multicalibration as supplementary sanity checks. These results are not the primary achievement of our model but help readers situate it against prior work. We report results on the subset of the data where our model makes predictions (omitting $d_{nf}$ nodes), which enables a direct comparison across all four models.

\subsubsection{Accuracy}

\begin{table}[ht]
\centering
\small
\begin{tabular}{l|rrrr}
\textbf{Dataset} & \textbf{FB} & \textbf{DT} & \textbf{NN} & \textbf{PMC} \\
\hline
Adult & 94.2\% & 93.5\% & 93.4\% & 92.6\% \\
COMPAS & 77.6\% & 67.4\% & 68.7\% & 68.5\% \\
Bank Marketing & 93.6\% & 91.0\% & 91.5\% & 69.1\% \\
\end{tabular}
\caption{Overall accuracy (omitting $d_{nf}$ nodes) on the full dataset.}
\label{tab:accuracy_summary}
\end{table}

Our model achieves the highest accuracy on all three datasets. The margin over DT and NN is modest on Adult and Bank Marketing (0.7 and 2.1--2.6 percentage points), but substantial on COMPAS (77.6\% vs 67.4\%/68.7\%), where the smaller sample size and more heterogeneous target structure make aggregation across overlapping $v$~nodes particularly useful. PMC's accuracy on Bank Marketing drops sharply (69.1\%), a direct consequence of its aggressive multiplicative corrections.

\subsubsection{Multiaccuracy}

Multiaccuracy measures a model's accuracy across protected subgroups, 
not just on aggregate~\cite{blum2018multiaccuracy}.
Table~\ref{tab:multiacc_summary} reports multiaccuracy for
the protected groups defined for each dataset (omitting $d_{nf}$
nodes). Our model achieves the highest accuracy in every cell
of every dataset. The margin is small on Adult (where all four
models are within roughly two percentage points), but widens
considerably on COMPAS (FB exceeds the strongest baseline by
$6$--$16$ percentage points across cells) and on the smaller
Bank Marketing groups (where PMC's multiplicative corrections
collapse calibration to such an extent that its per-cell accuracy
falls to as low as $26.7\%$).

\begin{table}[ht]
\centering
\small
\begin{tabular}{ll|rrrr}
\textbf{Group} & & \textbf{FB} & \textbf{DT} & \textbf{NN} & \textbf{PMC} \\
\hline
\multicolumn{6}{l}{\emph{Adult}} \\
Non-white & Female & 95.8 & 95.3 & 95.5 & 94.4 \\
Non-white & Male   & 90.8 & 89.5 & 89.2 & 88.0 \\
White     & Female & 96.3 & 95.8 & 95.6 & 95.6 \\
White     & Male   & 93.0 & 92.3 & 92.2 & 90.9 \\
\hline
\multicolumn{6}{l}{\emph{COMPAS}} \\
Caucasian     & Female & 84.6 & 68.5 & 71.0 & 68.0 \\
Caucasian     & Male   & 77.6 & 66.9 & 68.7 & 68.5 \\
Non-Caucasian & Female & 79.2 & 69.3 & 68.9 & 69.3 \\
Non-Caucasian & Male   & 75.9 & 67.0 & 68.3 & 68.4 \\
\hline
\multicolumn{6}{l}{\emph{Bank Marketing}} \\
Divorced & 25--60        & 93.8 & 91.0 & 91.9 & 82.6 \\
Divorced & ${<}25$ or ${>}60$ & 90.5 & 72.7 & 73.4 & 38.8 \\
Married  & 25--60        & 95.0 & 93.5 & 93.8 & 81.6 \\
Married  & ${<}25$ or ${>}60$ & 86.0 & 74.4 & 74.8 & 31.8 \\
Single   & 25--60        & 92.0 & 89.6 & 90.0 & 46.1 \\
Single   & ${<}25$ or ${>}60$ & 89.8 & 80.2 & 80.6 & 26.7 \\
\end{tabular}
\caption{Multiaccuracy: per-cell accuracy (\%) on each dataset's protected groups, omitting $d_{nf}$ nodes. Higher is better. Our model dominates every cell on every dataset.}
\label{tab:multiacc_summary}
\end{table}

\subsubsection{Multicalibration}

Table~\ref{tab:multical_adult} shows multicalibration results on
Adult (omitting $d_{nf}$ nodes). On Adult, all four models are
well calibrated, with PMC achieving the smallest errors as
expected given its explicit multicalibration objective and our
model close behind. The pattern is more nuanced on the other
two datasets (full tables in Appendix~\ref{app:multical_extra}):
on COMPAS our model has the smallest calibration errors of any
model, and on Bank Marketing PMC's multiplicative corrections
inflate calibration error sharply on smaller cells while our
model and the neural network remain well calibrated. Across the
three datasets, our model matches or beats DT and NN on every
cell, and matches or beats PMC on COMPAS and Bank Marketing.

\begin{table}[ht]
\centering
\small
\begin{tabular}{l|rrrr}
\textbf{Group} & \textbf{FB} & \textbf{DT} & \textbf{NN} & \textbf{PMC} \\
\hline
Female non-white & $-0.025$ & $-0.043$ & $-0.038$ & $-0.008$ \\
Male non-white   & $-0.048$ & $-0.069$ & $-0.058$ & $-0.013$ \\
Female white     & $-0.015$ & $-0.017$ & $-0.013$ & $-0.006$ \\
Male white       & $-0.032$ & $-0.036$ & $-0.031$ & $\phantom{-}0.003$ \\
\end{tabular}
\caption{Multicalibration error (predicted rate $-$ observed rate) on Adult, omitting $d_{nf}$ nodes. Values closer to zero indicate better calibration.}
\label{tab:multical_adult}
\end{table}

Crucially, PMC's strong multicalibration does not prevent it from violating $d$ and $v$~node consistency in several cases, as demonstrated in the previous sections. This illustrates a central finding: satisfying a standard fairness metric does not guarantee that individual predictions are statistically consistent with the observed evidence.

\section{Discussion}
\label{sec:discussion}

The results above can be understood as consequences of a single
underlying difference: standard ML models reason from features,
while our framework reasons from the groups that an individual
belongs to. This distinction has consequences along four connected
dimensions, all of which trace back to the difference between
frequentist and Bayesian inference.

\subsection{Reasoning from groups rather than features}

A standard ML model operates by learning associations between
input features and the target variable across the training data.
Predictions are made by evaluating these learned associations at
the individual's feature values, with the individual instance as
the unit of reasoning.
Our framework operates differently.
For a given individual, prediction is informed by two kinds of
group membership simultaneously.
First, their $d$~node: the set of all individuals in the data
who share identical values on every input feature.
The empirical target distribution of this group is read directly
from the data, and a prediction is made only if it is
statistically consistent with that distribution.
Second, their $v$~nodes: the broader subgroups defined by partial
feature specifications to which the individual also belongs.
Each $v$~node places an additional constraint on the prediction,
ensuring it is also consistent with the aggregate evidence across
overlapping subgroups.
This layered structure means that a prediction must simultaneously
respect the evidence at the most granular level (the $d$~node)
and at every relevant subgroup level (the $v$~nodes), rather than
averaging across the population as a whole.

\subsection{The importance of sample size}

This approach makes sample size central in a way that frequentist
models do not.
A frequentist model treats a 7/10 observed rate in a $d$~node
the same as a 700/1000 rate: both yield the same point estimate
and the model predicts accordingly.
In the Bayesian framework, these two cases are treated very
differently.
The posterior is much wider for the smaller node, and the
$V_{\min}/V_{\max}$ bounds widen correspondingly, meaning a
prediction is withheld unless the sample is large enough to
support a unique consistent outcome.
The same logic applies at the $v$~node level: a constraint derived
from a small subgroup carries less evidential weight and produces
wider bounds than one derived from a large subgroup.

This pattern is visible in the experimental results. On COMPAS, the
smallest of the three datasets and the one with the most
heterogeneous target structure, our accuracy advantage over the
baselines is substantial (roughly ten percentage points,
Table~\ref{tab:accuracy_summary}); on the larger Adult and Bank
Marketing datasets the margin is considerably smaller. Where sample
sizes are smaller and subgroup structure is more influential, the
Bayesian posterior's principled treatment of sample size gives the
framework its clearest advantage over frequentist alternatives.

\subsection{Structural protection against overfitting}
\label{sec:overfitting}

The concept of overfitting arises naturally from the frequentist
perspective on this problem.
Because a frequentist model treats observed patterns as directly
generalisable, it can learn associations that reflect the training
sample rather than the underlying population.
The standard response, regularisation, penalises model complexity
to discourage this, but at a cost: it discards information that
tends to be concentrated in small groups with limited
observations.
Our framework does not overfit in this sense because it makes
no claim beyond what the directly observed group-level evidence
supports.
It makes a prediction only when the direct evidence from the
relevant groups is sufficient to support one, and it abstains
otherwise.
The result is not a complete set of predictions, but every
prediction that is made is statistically defensible.
This is why our evaluation protocol presents the framework with
the full dataset rather than a training subset: more data
sharpens the posteriors without risk of overfitting, and the
complete $d$/$v$~node structure is needed to assess consistency.

\subsection{Implications for fairness}

The fairness implications follow directly from this.
Minority demographic groups are disproportionately represented
among small $d$~nodes, and small $d$~nodes are precisely where
frequentist and Bayesian inference diverge most sharply.
A frequentist model may produce a confident prediction for a
small minority subgroup on the basis of limited evidence that
would not survive Bayesian scrutiny at the $d$ or $v$~node level.
Regularisation compounds this: the information it discards is
concentrated in small nodes, which are disproportionately those
belonging to minority groups.
By tying prediction eligibility to the statistical weight of the
evidence at each level of the node hierarchy, rather than to a
global loss function, our framework gives small groups the same
inferential standard as large ones.

\subsection{Abstention as a diagnostic tool}

The abstention mechanism also serves as a diagnostic.
When a $d$~node is labelled $d_{nf}$, it signals that no
deterministic prediction is consistent with the observed data at
the chosen confidence level $\alpha$: the sample contains
individuals both with and without the target, and is not large
enough to resolve the ambiguity.
This identifies exactly which segments of the population require
additional variables or more data before a consistent
prediction becomes possible, providing actionable guidance for
data collection.
In practice, $d_{nf}$ cases can be routed to human reviewers
while the algorithm handles the remaining cases, a division of
labour that aligns with the transparency and human oversight
requirements of recent regulation such as the EU AI
Act~\cite{euaiact}.
On the datasets examined here, this amounts to roughly 10\% of
individuals on COMPAS and 11\% on Bank Marketing requiring human
review, with Adult considerably higher at 50\% (suggesting the
available variables are insufficient for reliable inference
across much of that population).

\subsection{Limitations}

Several limitations remain.
The framework requires categorical inputs and a binary target
(a multi-class extension using the Dirichlet distribution is
feasible but not explored here).
Discretising continuous variables introduces subjective thresholds
that may affect prediction quality.
When the training data itself encodes historical or structural
disadvantage, the question of how to draw fair inferences from
unfair evidence remains open; the prior provides one mechanism
for intervention, but its practical implications require further
work.
Computationally, the Gurobi MIP handles the datasets studied here
comfortably on commodity hardware (under two minutes for Bank
Marketing, the largest, with around 140{,}000 $v$~nodes); datasets
with substantially more features or finer discretisations may
require further engineering to keep the constraint system
tractable.
Although the constraints can be satisfied by more than one assignment,
the $v$~node likelihood selector makes a clear choice: at $\alpha^*$
the likelihood ratio between the selected solution and the runner-up is
$3.12\times 10^5$ on COMPAS, $2.40\times 10^{15}$ on Bank Marketing,
and $5.08\times 10^{15}$ on Adult.
Finally, as noted in Section~\ref{sec:unseen}, the procedure
for handling unseen $d$~nodes requires re-solving the constraint
system per instance, limiting the framework to batch settings.
A complementary empirical evaluation on a held-out portion of
each dataset (alongside the full-dataset consistency analysis
presented here) is left for future work.

\section{Conclusion}
\label{sec:conclusion}

This paper has presented a framework for statistically consistent
prediction grounded in Bayesian inference. We defined two
requirements for consistency (determinism and statistical
consistency with the inferred target distribution) and showed that
these requirements, when enforced exhaustively across all subgroups
in a categorical dataset, give rise to a classifier with three
properties not offered by existing approaches: predictions that are
guaranteed to be consistent with the observed evidence for every
group and subgroup, principled abstention when no consistent
deterministic prediction exists, and multicalibration competitive
with models explicitly optimised for it, as a by-product of the
consistency guarantee rather than as a direct objective.

Our evaluation on three benchmark datasets demonstrated that
standard ML models, including a decision tree, a neural network,
and a proportional multicalibration post-processor explicitly designed for
fairness, all produce predictions that are statistically
inconsistent with the observed data for a significant proportion
of subgroups. The Fair Bayesian classifier achieves zero
consistency error by construction and exceeds baseline accuracy
on every dataset tested. The results also demonstrate that
satisfying a standard fairness metric such as multicalibration
does not prevent a model from making predictions that contradict
the statistical evidence at the subgroup level. These findings
have direct implications for algorithmic fairness: minority
demographics are disproportionately represented among small
$d$~nodes, and it is precisely in these small nodes that
regularisation discards the most information and where our
framework diverges most sharply from standard approaches.

Several directions for future work follow from these results.
The most immediate is extension of the framework to continuous
variables and regression tasks, which would require different
distributional assumptions and inference procedures. The question 
of how to choose priors when
the training data encodes historical or structural disadvantage
also remains open: the prior offers a principled mechanism for
intervention, but its practical calibration requires dedicated
study. Finally, scaling the algorithm to datasets with
substantially more features or finer discretisations (where the
constraint pool grows accordingly) will require targeted
engineering, whether through parallelisation of the unseen-node
procedure or specialised solver techniques.

Taken together, these contributions offer a path toward prediction
systems whose individual outputs can be inspected, justified, and
routed to human review when the evidence does not support a
decision, properties increasingly required of high-stakes
algorithmic systems by regulation such as the EU AI
Act~\cite{euaiact}.

\section{Acknowledgments}

This work was funded by Taighde Éireann -- Research Ireland through the
Research Ireland Centre for Research Training in Machine Learning
(18/CRT/6183).

\printbibliography

\appendix
\section{Pseudocode}
\label{app:pseudocode}

\subsubsection{Data Preparation (Steps 1--5)}

The first step is to set the parameters of the algorithm:

\begin{itemize}
    \item $\alpha$: Significance level. Larger values make the requirements for statistical consistency more stringent. We use $10^{-5}$ as the starting point of the binary search over $\alpha$ described in \S\ref{sec:alpha}; the reported value of $\alpha^*$ is the loosest level at which the constraint system remains feasible.
    \item \texttt{intervals}: Resolution for probability density approximation. We use a default value of 1000.
    \item $a_0$ and $b_0$: Priors. We take the uniform prior of $a_0 = b_0 = 1$ as our default.
    \item \texttt{hbaseline}: Heterogeneity baseline. $d$ or $v$~nodes larger than \texttt{hbaseline} will be scaled to be of size \texttt{hbaseline}. We use a default value of 100.
\end{itemize}

Steps 1--5 are contained within the function \texttt{DataPrep}:
\begin{algorithm}[ht]
\caption{DataPrep($\text{data}, a_0, b_0$)}
\label{alg:dataprep}
\KwIn{Data, prior parameters $a_0$, $b_0$}
\KwResult{Results in \texttt{dndf} ($d$~node DataFrame)}

Summarise data into a list of every combination of variable values\;
Store each combination with its count and target count in \texttt{combos}\;

\For{$i \in \{1, \ldots, \text{len(combos)}\}$}{
    Compute Beta density using data and priors to resolution \texttt{xvalues}\;
    Assign density to \texttt{combos[`m'][$i$]}\;
}

\For{$i \in \{1, \ldots, \text{len(combos)}\}$}{
    Compute Beta-Binomial density using Binomial PMF and \texttt{m} to resolution \texttt{xvalues}\;
    Assign density to \texttt{combos[`M'][$i$]}\;
}

\For{$i \in \{1, \ldots, \text{len(combos)}\}$}{
    Compute CDF of distribution stored in \texttt{M}\;
    Assign $\arg\max$ of \texttt{combos[`cdf\_M'][$i$]} $> \alpha$ to \texttt{combos[$V_{\min}$][$i$]}\;
    Assign $\arg\max$ of \texttt{combos[`cdf\_M'][$i$]} $> 1-\alpha$ to \texttt{combos[$V_{\max}$][$i$]}\;
}

Sort each $d$~node into a $d$ category\;
Assign predictions for $d_0$ and $d_1$ nodes\;
\Return{\texttt{dndf}}
\end{algorithm}

The output of this function is a list of $d$~nodes, each with an inferred Beta distribution, corresponding Beta-Binomial distribution, $d$ category and prediction (in the case of $d_0$ and $d_1$ nodes).

\subsubsection{Generating \texorpdfstring{$v$}{v}~nodes (Step 6)}

The function \texttt{GenVs} generates the list of $v$~nodes from our list of $d$~nodes:

\begin{algorithm}[ht]
\caption{GenVs(data, dndf, $\alpha$, protected\_vars, intervals)}
\label{alg:genvs}
\KwIn{Data, \texttt{dndf}, significance level $\alpha$, protected variables, intervals}
\KwResult{Updated \texttt{vndf}}

\For{$d$ in \texttt{dndf}}{
    Generate all possible $v$~nodes that could contain $d$ and store in \texttt{vndf}\;
}
\For{$v$ in \texttt{vndf}}{
    Create a copy of \texttt{dndf} and filter it to the $d$ children of $v$\;
    Calculate and store the count, target count and $d$ children nodes of $v$\;
}
\For{$v$ in \texttt{vndf}}{
    Create a copy of \texttt{vndf} and filter it to $v$~nodes identical to $v$ but with one more ambiguous column\;
    Store the indices as the direct $v$ children of $v$\;
}
\Return{\texttt{vndf}}
\end{algorithm}

Rather than generating all permutations of variables (which for e.g. ten variables with three categories each would yield $(3+1)^{10} = 1{,}048{,}576$ candidates), we generate $v$~nodes based on the list of $d$~nodes. This prevents us having to consider $v$~nodes that aren't present in the data. After determining the count, target count, $d$~node children and direct $v$~node children, we reduce the list further. First, we remove all $v$~nodes that only have a single $d$~node as a child, as the resulting constraint will be identical to that of the $d$~node. Next, we drop all but one of nodes that have the same $d$ children as they will all contain the same information. We also drop $v$~nodes where $V_{\min} = 0$ and $V_{\max} = \text{Count}$ as their resulting constraints will always be satisfied.

These steps reduce the number of constraints significantly. Table~\ref{tab:vnode_reduction} shows the reduction in the number of $v$~nodes for each dataset.

\begin{table}[ht]
\centering
\begin{tabular}{lcc}
\hline
Dataset & Initial No. & Final No. \\
\hline
COMPAS & 380,160 & 32,483 \\
Adult & 1,409,024 & 81,380 \\
Bank Marketing & 2,269,184 & 142,030 \\
\hline
\end{tabular}
\caption{Reduction in number of $v$~nodes for each dataset.}
\label{tab:vnode_reduction}
\end{table}

\subsubsection{Computing \texorpdfstring{$v$}{v}~node Distributions (Step 7)}

Calculating the Beta and Beta-Binomial approximations for each $v$~node is done by the function \texttt{CalcMs}. The function \texttt{SplittingSolver} is a recursive function used within \texttt{CalcMs} to handle $v$~nodes with more than two children.
\begin{algorithm}[ht]
\caption{CalcMs(data, dndf, vndf, hbaseline)}
\label{alg:calcms}
\KwIn{\texttt{data}, \texttt{dndf}, \texttt{vndf}, heterogeneity baseline \texttt{hbaseline}}
\KwResult{Updated \texttt{vndf} with computed densities}

Begin calculations with $v$~nodes with a single ambiguous variable, then two and so on\;

\For{$l$ in range(lencols)}{
    \For{$v$ in \texttt{vndf} where the number of ambiguous variables equals $l$}{
        Check if $v$ has any direct $v$ children nodes and if so check for a \emph{complete split}\;
        \eIf{a complete split exists}{
            Use those $v$~nodes as the direct children (\texttt{dcs}) of $v$\;
        }{
            Use the $d$ children of $v$ as \texttt{dcs}\;
        }
        \eIf{$|\texttt{dcs}| \leq 2$}{
            Let $\text{pta} = \texttt{dcs[`m'][0]}$, $\text{ptb} = \texttt{dcs[`m'][1]}$\;
            Calculate $p_{A|U}$ using the Beta CDF and each child's count\;
            Generate samples $x, y, z$ from densities \texttt{pta}, \texttt{ptb}, and $p_{A|U}$\;
            Let $w = x \cdot z + y \cdot (1-z)$\;
            Set \texttt{vndf[`m'][$i$]} to normalised histogram of $w$\;
        }{
            Set \texttt{vndf[`m'][$i$]} to result of \texttt{SplittingSolver}\;
        }
    }
}
\For{$v$ in \texttt{vndf}}{
    \If{$v$ is larger than \texttt{hbaseline}}{
        Scale Beta distribution to account for heterogeneity\;
    }
    Calculate Beta-Binomial approximation for $v$\;
}
\Return{\texttt{vndf}}
\end{algorithm}

\begin{algorithm}[ht]
\caption{SplittingSolver(ch\_data)}
\label{alg:splitting}
\KwIn{List of child distributions \texttt{ch\_data}}
\KwResult{Single merged density}

\If{$|\texttt{ch\_data}| = 1$}{
    \Return{\texttt{ch\_data}}\;
}
Divide \texttt{ch\_data} into pairs\;
\For{each pair in pairs}{
    Calculate density as in \texttt{CalcMs}\;
    Merge pair into a single entry with new density\;
}
Store new, shorter list of distributions as \texttt{ch\_data}\;
\Return{\texttt{SplittingSolver(ch\_data)}}
\end{algorithm}

This function makes use of several key aspects of our $v$~node structure to maximise efficiency. The inferred Beta distribution approximation for a given $v$~node can be calculated directly from the $d$ children nodes, but this may not be the most efficient approach. Large $v$~nodes with many ambiguous variables may have hundreds of $d$ children nodes; combining their distributions would be computationally infeasible. Instead, we begin at the `lowest level' of $v$~nodes, those with only a single ambiguous variable. Their distributions must be calculated from their $d$ children nodes, but these calculations will all be reasonably simple. Then, we move to the next level up and look at $v$~nodes with two ambiguous variables. Rather than using the distributions of their $d$ children nodes, we use the distributions of their direct $v$ children nodes (i.e.\ the level one $v$~nodes completely contained within the level two $v$~nodes). This drastically reduces the number of computations needed and can be repeated as we move up through the $v$~nodes.

When using direct $v$ children nodes in this way, we must ensure that the children nodes make up a `complete split', i.e.\ that the children nodes contain every individual in the parent node between them. If there are multiple groups of $v$ child nodes that make up a complete split, we choose the group with the fewest number of $d$~nodes as it will be the least expensive to calculate. If no complete split of $v$ children nodes exists for a given $v$~node, we simply use the $d$ children nodes.

Once the children have been selected, we approximate the $v$~node's Beta distribution using sampling. The process is simplest when there are only two children as we can simply apply the method described in Section~\ref{sec:vnode_distributions}. When there are more than two children, we must calculate intermediate approximations in a pairwise fashion until a single distribution remains. This is achieved using the recursive function \texttt{SplittingSolver}. After calculating the Beta approximation, we calculate the Beta-Binomial approximation (factoring in heterogeneity for large $d$ or $v$~nodes by scaling).

\subsubsection{Constraint Satisfaction (Steps 8--12)}

Once each $v$~node has a Beta-Binomial approximation, we
read off its $V_{\min}$ and $V_{\max}$ boundaries (step 8) in
the same way as for the $d$~nodes. Before solving, we
identify any $v$~node that is mostly composed of $d_{nf}$
(abstaining) children (step 9) and widen its boundaries to
$[0, N_v]$, so those nodes cannot block an otherwise valid
solution.

The constraints are then handed to the Gurobi Optimiser:

\begin{algorithm}[ht]
\caption{Gurobi Process}
\label{alg:gurobi}
\KwResult{Solved predictions for $d_{am}$ nodes}

Initialise the Gurobi model\;
Add binary variables $x[i]$ to the model, each representing a $d$~node prediction\;
Assign predictions for $d_0$ and $d_1$ nodes\;
Add constraints based on $V_{\min}$ and $V_{\max}$ bounds from \texttt{vndf} and \texttt{dndf}\;
Set the objective as in Equation~\ref{eq:objective_full}\;
Configure Gurobi to return a pool of the top $K$ feasible solutions\;
Run the Gurobi model and retrieve the solution pool\;
Re-score each pool member by the $v$-node log-likelihood $L(\mathbf{x})$ (Equation~\ref{eq:vnode_ll})\;
Return the pool member maximising $L(\mathbf{x})$\;
\end{algorithm}

Each $d$~node is represented by a binary variable $x_d$,
equal to $1$ if we predict the positive label and $0$
otherwise. We fix $x_d$ in advance for the $\mathcal{D}_0$
and $\mathcal{D}_1$ nodes (whose labels are already
decided), leaving the MIP to solve for the remaining nodes
in $\mathcal{D}_{am} \cup \mathcal{D}_{nf}$
(Section~\ref{sec:d_categories}). For each $v$~node, a
constraint requires the count-weighted sum of its $d$
children's predictions (using the $d$~node sizes $N_d$ as
coefficients) to fall inside the $v$~node's
$V_{\min}/V_{\max}$ window. The MIP then ranks feasible
assignments using the objective
\begin{equation}
\label{eq:objective_full}
\max \;\; \sum_{d \in \mathcal{D}_{am} \cup \mathcal{D}_{nf}} w_d \cdot x_d, \qquad w_d \;=\; \log\!\frac{\hat{p}_d}{1 - \hat{p}_d} \cdot \log(N_d + 1)
\end{equation}
where $\hat{p}_d = (T_d + 1)/(N_d + 2)$ is the node's
posterior mean positive rate under the uniform prior
(Equation~\ref{eq:posterior_uniform}). The weight $w_d$ is
positive when the $d$~node has seen more positives than
negatives ($\hat{p}_d > 1/2$) and negative otherwise; its
magnitude grows with the strength of the evidence and, via
the $\log(N_d + 1)$ factor, with the size of the node.

Without the $v$~node constraints, each $d$~node would be
decided on its own: the MIP would set $x_d = 1$ for every
node with $w_d > 0$, i.e.\ for every node whose data favour
the positive label. The $v$~node constraints tie these
decisions together, and can rule out that simple choice
when the labels preferred by several $d$~nodes would
together fall outside the $V_{\min}/V_{\max}$ window of
their parent $v$~node. The MIP then picks the feasible
assignment of largest total weight, which flips labels at
$d$~nodes where the evidence is weak (small $|w_d|$) in
preference to those where it is strong. The constraints
must hold regardless of the objective; $w_d$ serves only to
rank feasible assignments.

The objective above uses only the $d$~node posteriors; the
$v$~node posteriors
(Section~\ref{sec:vnode_distributions}) enter the MIP only
as feasibility constraints. To also use them at selection
time, we ask Gurobi to return a pool of the $K$ best
feasible solutions under the $d$-objective
(\texttt{PoolSearchMode}$=2$, $K = 100$ in our experiments),
giving a pool
$\mathcal{P} = \{\mathbf{x}^{(1)}, \ldots, \mathbf{x}^{(|\mathcal{P}|)}\}$.

Each pool member is then scored by how well it agrees with
the $v$~node distributions. For a $v$~node $v$ with
Beta-Binomial PMF $M_v$
(Section~\ref{sec:vnode_distributions}) and $d$-children
$\mathrm{dc}(v)$, the count of observed positives inside
$v$ that an assignment $\mathbf{x}$ labels positive is
\begin{equation}
\label{eq:vnode_tv}
t_v(\mathbf{x}) \;=\; \sum_{d \in \mathrm{dc}(v)} T_d \, x_d,
\end{equation}
and the assignment's overall score is the sum of the
log-probabilities of these counts under the $M_v$:
\begin{equation}
\label{eq:vnode_ll}
L(\mathbf{x}) \;=\; \sum_{v \in \mathcal{V}} \log M_v\!\big(t_v(\mathbf{x})\big).
\end{equation}
Log-probabilities are pre-computed and
floored at $\log(10^{-300})$ to avoid $\log 0$. The final
prediction is the pool member with the highest $L$
(Equation~\ref{eq:selector}). This lets us pick between
pool members using the richer $v$~node evidence, which
cannot be expressed as a linear MIP objective. A final
check confirms all boundaries hold for the chosen
solution.

\section{Baseline Hyperparameter Configurations}
\label{app:hyperparams}

This appendix reports the hyperparameter search grids
and the final configuration selected by ten-fold
stratified cross-validation for each baseline model
and dataset.
All models are implemented in scikit-learn~\cite{sklearn}.

\subsection*{Decision Tree (\texttt{DecisionTreeClassifier})}

\begin{table}[ht]
\centering
\small
\begin{tabular}{l|l}
\textbf{Parameter} & \textbf{Values searched} \\
\hline
\texttt{criterion}           & gini, entropy \\
\texttt{max\_depth}          & None, 5, 10, 15 \\
\texttt{min\_samples\_split}  & 2, 5, 10 \\
\texttt{min\_samples\_leaf}   & 1, 2, 5 \\
\texttt{max\_features}       & None, sqrt, log2 \\
\texttt{ccp\_alpha}          & 0.0, 0.001, 0.01, 0.1 \\
\end{tabular}
\caption{Decision Tree search grid (864 candidates, 10-fold CV).}
\label{tab:dt_grid}
\end{table}

The final selected configurations for each dataset
are reported in Table~\ref{tab:dt_selected}.

\begin{table}[ht]
\centering
\small
\begin{tabular}{l|llllll}
\textbf{Dataset} & \textbf{criterion} & \textbf{depth}
  & \textbf{max\_feat.} & \textbf{split} & \textbf{leaf} & \textbf{ccp} \\
\hline
Adult        & entropy & 5    & None & 2 & 1 & 0.0  \\
COMPAS       & entropy & None & None & 2 & 1 & 0.01 \\
Bank Mktg.   & gini    & 5    & None & 2 & 5 & 0.0  \\
\end{tabular}
\caption{Decision Tree: final selected configuration per dataset.}
\label{tab:dt_selected}
\end{table}

\subsection*{Neural Network (\texttt{MLPClassifier})}

\begin{table}[ht]
\centering
\small
\begin{tabular}{l|l}
\textbf{Parameter} & \textbf{Values searched} \\
\hline
\texttt{hidden\_layer\_sizes} & (10,10), (100,10), (50,50) \\
\texttt{activation}          & relu, logistic \\
\texttt{solver}              & adam \\
\texttt{learning\_rate}      & adaptive \\
\texttt{alpha} ($\ell_2$)    & $10^{-4}$, $10^{-3}$ \\
\texttt{max\_iter}           & 200 \\
\end{tabular}
\caption{Neural Network search grid (12 candidates, 10-fold CV).}
\label{tab:nn_grid}
\end{table}

The final selected configurations for each dataset
are reported in Table~\ref{tab:nn_selected}.
All use the \texttt{adam} solver with an
\texttt{adaptive} learning-rate schedule.

\begin{table}[ht]
\centering
\small
\begin{tabular}{l|llll}
\textbf{Dataset} & \textbf{hidden\_layers}
  & \textbf{activation} & \textbf{alpha} \\
\hline
Adult        & $(10,10)$ & logistic & $10^{-3}$ \\
COMPAS       & $(50,50)$ & logistic & $10^{-3}$ \\
Bank Mktg.   & $(50,50)$ & logistic & $10^{-3}$ \\
\end{tabular}
\caption{Neural Network: final selected configuration per dataset.}
\label{tab:nn_selected}
\end{table}

\subsection*{Proportional Multi-Calibrator (PMC)}

The PMC post-processor is applied to a logistic regression
base learner (\texttt{LogisticRegression}, $C = 1.0$,
\textit{max\_iter}~=~1000).
The \texttt{MultiCalibrator} auditor is initialised
with the protected attributes for each dataset;
$n\_\text{bins} = 5$; all other parameters use the
library defaults from
\texttt{mcboost}~\cite{lacava2023pmc}.
No cross-validation is applied to PMC: the base
learner is fitted on the training set and the
calibrator is applied as a post-processing step.
Binary predictions are produced by thresholding
calibrated probabilities at the dataset base rate.

\section{Alpha Sensitivity: Additional Datasets}
\label{app:alpha_additional}

Tables~\ref{tab:alpha_sweep_compas} and~\ref{tab:alpha_sweep_bm} replicate the alpha sensitivity analysis of Section~\ref{sec:alpha} on the COMPAS and Bank Marketing datasets, respectively. In both cases the same qualitative pattern holds: the system is infeasible for sufficiently large $\alpha$, a feasibility threshold $\alpha^*$ is crossed at which a consistent solution first exists, and the number of active constraints decreases monotonically as $\alpha$ is further relaxed. The optimal solution is unique (N@Best = 1) at every feasible $\alpha$, confirming that $\alpha^*$ produces a stable, well-identified prediction assignment on these datasets as well.

\begin{table}[ht]
\centering
\small
\begin{tabular}{r|ccc}
$\alpha$ & \textbf{Feasible} & \textbf{Active Constraints} & \textbf{Unique (N@Best)} \\
\hline
$10^{-3}$               & No  & ---    & --- \\
$10^{-4}$               & No  & ---    & --- \\
$10^{-5}$               & No  & ---    & --- \\
$5 \times 10^{-6}$      & No  & ---    & --- \\
$2 \times 10^{-6}$      & Yes & 11,291 & 1   \\
$10^{-6}$               & Yes & 10,792 & 1   \\
$10^{-7}$               & Yes &  9,446 & 1   \\
\end{tabular}
\caption{Alpha sensitivity analysis on the COMPAS dataset. Active constraints is the number of $v$~node constraints after pruning trivially satisfied bounds. N@Best = 1 indicates the optimal solution is unique at every feasible $\alpha$.}
\label{tab:alpha_sweep_compas}
\end{table}

\begin{table}[ht]
\centering
\small
\begin{tabular}{r|ccc}
$\alpha$ & \textbf{Feasible} & \textbf{Active Constraints} & \textbf{Unique (N@Best)} \\
\hline
$10^{-3}$               & No  & ---    & --- \\
$10^{-4}$               & No  & ---    & --- \\
$10^{-5}$               & No  & ---    & --- \\
$5 \times 10^{-6}$      & No  & ---    & --- \\
$2 \times 10^{-6}$      & Yes & 70,001 & 1   \\
$10^{-6}$               & Yes & 67,898 & 1   \\
$10^{-7}$               & Yes & 62,309 & 1   \\
\end{tabular}
\caption{Alpha sensitivity analysis on the Bank Marketing dataset. Active constraints is the number of $v$~node constraints after pruning trivially satisfied bounds. N@Best = 1 indicates the optimal solution is unique at every feasible $\alpha$.}
\label{tab:alpha_sweep_bm}
\end{table}

\section{Multicalibration: Additional Datasets}
\label{app:multical_extra}

This appendix reports the multicalibration tables for COMPAS
(Table~\ref{tab:multical_compas}) and Bank Marketing
(Table~\ref{tab:multical_bm}), complementing the Adult table
(Table~\ref{tab:multical_adult}) in Section~\ref{sec:performance}.
Values closer to zero indicate better calibration.

\begin{table}[ht]
\centering
\small
\begin{tabular}{ll|rrrr}
\textbf{Group} & & \textbf{FB} & \textbf{DT} & \textbf{NN} & \textbf{PMC} \\
\hline
Caucasian     & Female & $\phantom{-}0.052$ & $-0.196$ & $-0.212$ & $-0.099$ \\
Caucasian     & Male   & $-0.014$ & $-0.153$ & $-0.094$ & $-0.046$ \\
Non-Caucasian & Female & $-0.021$ & $-0.121$ & $-0.148$ & $-0.035$ \\
Non-Caucasian & Male   & $\phantom{-}0.042$ & $-0.080$ & $\phantom{-}0.009$ & $\phantom{-}0.041$ \\
\end{tabular}
\caption{Multicalibration error (predicted rate $-$ observed rate) on COMPAS, omitting $d_{nf}$ nodes.}
\label{tab:multical_compas}
\end{table}

\begin{table}[ht]
\centering
\small
\begin{tabular}{ll|rrrr}
\textbf{Group} & & \textbf{FB} & \textbf{DT} & \textbf{NN} & \textbf{PMC} \\
\hline
Divorced & 25--60               & $-0.017$ & $-0.004$ & $-0.003$ & $\phantom{-}0.107$ \\
Divorced & ${<}25$ or ${>}60$    & $\phantom{-}0.043$ & $-0.187$ & $\phantom{-}0.030$ & $\phantom{-}0.612$ \\
Married  & 25--60               & $-0.016$ & $-0.050$ & $-0.030$ & $\phantom{-}0.137$ \\
Married  & ${<}25$ or ${>}60$    & $\phantom{-}0.012$ & $-0.174$ & $-0.062$ & $\phantom{-}0.682$ \\
Single   & 25--60               & $-0.019$ & $-0.034$ & $-0.018$ & $\phantom{-}0.527$ \\
Single   & ${<}25$ or ${>}60$    & $-0.020$ & $-0.136$ & $-0.073$ & $\phantom{-}0.733$ \\
\end{tabular}
\caption{Multicalibration error (predicted rate $-$ observed rate) on Bank Marketing, omitting $d_{nf}$ nodes. PMC's multiplicative correction inflates calibration error sharply on smaller cells.}
\label{tab:multical_bm}
\end{table}

\section{Unseen Data Procedure}
\label{app:unseen}

This appendix gives a complete description of the
procedure used to handle test instances belonging to
$d$~nodes not seen during training.
It is included for completeness; extending this mechanism
to a general deployment setting is left as future work.

Let $D_{\text{train}}$ denote the original dataset and
$\hat{\theta}$ the solution obtained from solving the
constraints on $D_{\text{train}}$.
Given a test instance $x^*$ belonging to an unseen
$d$~node, we proceed as follows:

\begin{enumerate}
    \item \textbf{Check with prediction $T = 0$.}
    Append $(x^*, 0)$ to $D_{\text{train}}$ and re-solve
    the optimisation problem. Record whether the
    constraints remain feasible and, if so, the predicted
    label $\hat{y}_0$ returned for $x^*$.

    \item \textbf{Check with prediction $T = 1$.}
    Repeat the step above with $(x^*, 1)$ to check
    feasibility and predicted label $\hat{y}_1$.

    \item \textbf{Decision rule.}
    \begin{itemize}
        \item If both extended problems are infeasible,
        abstain from predicting.
        \item If exactly one is feasible, predict its label.
        \item If both are feasible and
        $\hat{y}_0 = \hat{y}_1$, predict the common label.
        \item If both are feasible and
        $\hat{y}_0 \neq \hat{y}_1$, abstain from predicting.
    \end{itemize}
\end{enumerate}

This procedure checks whether the existing evidence from
the $v$~nodes supports a unique, statistically consistent
prediction for the new individual, and abstains where it
does not. Instances that must be abstained from in this
way are referred to as unseen ambiguous nodes ($d_{ua}$).
The procedure requires re-solving the constraint system
for each unseen instance; parallelisation of the
independent re-solves is a natural avenue for improvement.

\section{Reproducibility Checklist for JAIR}

Select the answers that apply to your research -- one per item. 

\subsection*{All articles:}

\begin{enumerate}
    \item All claims investigated in this work are clearly stated.
    \textbf{[yes]}
    \item Clear explanations are given how the work reported substantiates the claims.
    \textbf{[yes]}
    \item Limitations or technical assumptions are stated clearly and explicitly.
    \textbf{[yes]}
    \item Conceptual outlines and/or pseudo-code descriptions of the AI methods introduced in this work are provided, and important implementation details are discussed.
    \textbf{[yes]}
    \item
    Motivation is provided for all design choices, including algorithms, implementation choices, parameters, data sets and experimental protocols beyond metrics.
    \textbf{[yes]}
\end{enumerate}

\subsection*{Articles containing theoretical contributions:}
Does this paper make theoretical contributions?
\textbf{[yes]}

If yes, please complete the list below.

\begin{enumerate}
    \item All assumptions and restrictions are stated clearly and formally.
    \textbf{[yes]}
    \item All novel claims are stated formally (e.g., in theorem statements).
    \textbf{[yes]}
    \item Proofs of all non-trivial claims are provided in sufficient detail to permit verification by readers with a reasonable degree of expertise (e.g., that expected from a PhD candidate in the same area of AI). \textbf{[yes]}
    \item
    Complex formalism, such as definitions or proofs, is motivated and explained clearly.
    \textbf{[yes]}
    \item
    The use of mathematical notation and formalism serves the purpose of enhancing clarity and precision; gratuitous use of mathematical formalism (i.e., use that does not enhance clarity or precision) is avoided.
    \textbf{[yes]}
    \item
    Appropriate citations are given for all non-trivial theoretical tools and techniques.
    \textbf{[yes]}
\end{enumerate}

\subsection*{Articles reporting on computational experiments:}
Does this paper include computational experiments? \textbf{[yes]}

If yes, please complete the list below.
\begin{enumerate}
    \item
    All source code required for conducting experiments is included in an online appendix
    or will be made publicly available upon publication of the paper.
    The online appendix follows best practices for source code readability and documentation as well as for long-term accessibility.
    \textbf{[yes]}
    \item The source code comes with a license that
    allows free usage for reproducibility purposes.
    \textbf{[yes]}
    \item The source code comes with a license that
    allows free usage for research purposes in general.
    \textbf{[yes]}
    \item
    Raw, unaggregated data from all experiments is included in an online appendix
    or will be made publicly available upon publication of the paper.
    The online appendix follows best practices for long-term accessibility.
    \textbf{[yes]}
    \item The unaggregated data comes with a license that
    allows free usage for reproducibility purposes.
    \textbf{[yes]}
    \item The unaggregated data comes with a license that
    allows free usage for research purposes in general.
    \textbf{[yes]}
    \item If an algorithm depends on randomness, then the method used for generating random numbers and for setting seeds is described in a way sufficient to allow replication of results.
    \textbf{[yes]}
    \item The execution environment for experiments, the computing infrastructure (hardware and software) used for running them, is described, including GPU/CPU makes and models; amount of memory (cache and RAM); make and version of operating system; names and versions of relevant software libraries and frameworks.
    \textbf{[yes]}
    \item
    The evaluation metrics used in experiments are clearly explained and their choice is explicitly motivated.
    \textbf{[yes]}
    \item
    The number of algorithm runs used to compute each result is reported.
    \textbf{[yes]}
    \item
    Reported results have not been ``cherry-picked'' by silently ignoring unsuccessful or unsatisfactory experiments.
    \textbf{[yes]}
    \item
    Analysis of results goes beyond single-dimensional summaries of performance (e.g., average, median) to include measures of variation, confidence, or other distributional information.
    \textbf{[yes]}
    \item
    All (hyper-) parameter settings for
    the algorithms/methods used in experiments have been reported, along with the rationale or method for determining them.
    \textbf{[yes]}
    \item
    The number and range of (hyper-) parameter settings explored prior to conducting final experiments have been indicated, along with the effort spent on (hyper-) parameter optimisation.
    \textbf{[yes]}
    \item
    Appropriately chosen statistical hypothesis tests are used to establish statistical significance
    in the presence of noise effects.
    \textbf{[NA]}
\end{enumerate}

\subsection*{Articles using data sets:}
Does this work rely on one or more data sets (possibly obtained from a benchmark generator or similar software artifact)?
\textbf{[yes]}

If yes, please complete the list below.
\begin{enumerate}
    \item
    All newly introduced data sets
    are included in an online appendix
    or will be made publicly available upon publication of the paper.
    The online appendix follows best practices for long-term accessibility with a license
    that allows free usage for research purposes.
    \textbf{[NA]}
    \item The newly introduced data set comes with a license that
    allows free usage for reproducibility purposes.
    \textbf{[NA]}
    \item The newly introduced data set comes with a license that
    allows free usage for research purposes in general.
    \textbf{[NA]}
    \item All data sets drawn from the literature or other public sources (potentially including authors' own previously published work) are accompanied by appropriate citations.
    \textbf{[yes]}
    \item All data sets drawn from the existing literature (potentially including authors’ own previously published work) are publicly available. \textbf{[yes]}
    \item All new data sets and data sets that are not publicly available are described in detail, including relevant statistics, the data collection process and annotation process if relevant.
    \textbf{[NA]}
    \item
    All methods used for preprocessing, augmenting, batching or splitting data sets (e.g., in the context of hold-out or cross-validation)
    are described in detail. \textbf{[yes]}
\end{enumerate}

\subsection*{Explanations on any of the answers above (optional):}

\textbf{Code and licensing.} The complete implementation is publicly
available at \url{https://github.com/owenon7/fairbayesian} under an MIT
licence.

\textbf{Data and licensing.} The three benchmark datasets used in this paper
are not redistributed in our repository (we link to them only): Adult and
Bank Marketing are released by the UCI Machine Learning Repository under the
Creative Commons Attribution 4.0 licence, and the COMPAS two-year recidivism
data is publicly available alongside ProPublica's 2016 investigation.
Citations for all three are given in Section~\ref{sec:experiments}, and
preprocessing steps (categorical binning, target renaming, removal of
high-cardinality columns) are described in Section~\ref{sec:experiments} and
implemented in the repository's \texttt{preprocess\_data.py}.

\textbf{Randomness and number of runs.} Three sources of randomness appear
in our pipeline. First, computing the $v$~node $M$-distributions involves a
Monte Carlo convolution step (5{,}000 samples per distribution, binned back
into the $K=1{,}000$-bin discretisation); this approximation is controlled by
a fixed NumPy seed of $0$. Second, baseline classifier training uses
\texttt{scikit-learn}'s pseudorandom number generator with a fixed seed of
$42$. Third, Gurobi's solution-pool sampling uses an MIP seed of $42$ with a
requested pool size of $100$. With these seeds fixed, every component is
exactly reproducible: rerunning the pipeline yields identical
$M$-distributions, baselines, and MIP solutions, as verified by a clean-room
reproduction from a fresh clone of the public repository. Each experiment
therefore consists of a single reproducible run; verification that the MIP
reaches a unique global optimum is performed by the
\texttt{verify\_uniqueness.py} script in the repository, whose results are
summarised in Section~\ref{sec:experiments}.

\textbf{Distributional information.} Beyond the headline accuracy and
consistency-error tables, we report node-level dataframes
(\texttt{dnodes.parquet}, \texttt{vnodes.parquet}) for every dataset that
allow readers to inspect the framework's predictions, posterior bounds, and
agreement with each baseline at the level of individual subgroups. We also
include an alpha-sensitivity sweep
(Table~\ref{tab:alpha_sweep}) showing how feasibility, the
selected solution, and the agreement between the two solution-selection
criteria evolve over five orders of magnitude in $\alpha$.

\textbf{Statistical hypothesis tests.} Marked as not applicable: with all
seeds fixed (see above), each dataset yields a single reproducible solution,
so there is no run-to-run sampling distribution against which to test the
reported numbers. 
\end{document}